\documentclass{article}

\usepackage[preprint]{neurips_2019}

\usepackage[utf8]{inputenc} 
\usepackage[T1]{fontenc}    
\usepackage{hyperref}       
\usepackage{url}            
\usepackage{booktabs}       
\usepackage{amsfonts}       
\usepackage{nicefrac}       
\usepackage{microtype}      
\usepackage{graphicx}
\usepackage{amsmath}
\usepackage{bbm}

\def \cN {\mathcal{N}}
\def \cS {\mathcal{S}}

\def \cX {\mathcal{X}}
\def \cP {\mathcal{P}}

\def \cF {\mathcal{F}}
\def \cI {\mathcal{I}}

\def \tr {{\rm tr}}

\def \EE {\mathbbm{E}}

\title{Degrees of Freedom Analysis of Unrolled Neural Networks}

\author{
  Morteza Mardani$^{1}$, Qingyun Sun$^{2}$, Vardan Papyan$^{2}$, Shreyas Vasanawala$^{3}$, \\ \textbf{John Pauly$^{1}$, and David Donoho$^{2}$} \\
  Depts. of Electrical Engineering$^{1}$, Statistics$^{2}$, and Radiology$^{3}$, Stanford University\\
  \texttt{\{morteza, qysun, papyan, vasanawala,pauly,donoho\}@stanford.edu} \\
}

\begin{document}





\maketitle




\begin{abstract}
%

Unrolled neural networks emerged recently as an effective model for learning inverse maps appearing in image restoration tasks. However, their generalization risk (i.e., test mean-squared-error) and its link to network design and train sample size remains mysterious. Leveraging the Stein's Unbiased Risk Estimator (SURE), this paper analyzes the generalization risk with its bias and variance components for recurrent unrolled networks. We particularly investigate the degrees-of-freedom (DOF) component of SURE, trace of the end-to-end network Jacobian, to quantify the prediction variance. We prove that DOF is well-approximated by the weighted \textit{path sparsity} of the network under incoherence conditions on the trained weights. Empirically, we examine the SURE components as a function of train sample size for both recurrent and non-recurrent (with many more parameters) unrolled networks. Our key observations indicate that: 1) DOF increases with train sample size and converges to the generalization risk for both recurrent and  non-recurrent schemes; 2) recurrent network converges significantly faster (with less train samples) compared with non-recurrent scheme, hence recurrence serves as a regularization for low sample size regimes.

\end{abstract}

\vspace{-2mm}

\section{Introduction}
\label{sec:introduction}
\vspace{-2mm}

Training deep neural networks typically demands abundant labeled data to achieve an acceptable generalization. Collecting valid labels, however, is costly if not impossible for certain applications such as medical imaging due to physical constraints and privacy concerns. This paper deals with imaging from compressive and noisy measurements, where labels are high-quality images that in medical imaging drive diagnostic decisions.


Outside the scarce-label setting, recent works adopts \textit{unrolled} neural networks to learn the inversion map for recovering an underlying image from compressive and corrupted measurements; see e.g., ~\cite{gregor2010learning,zhang2017beyond,chan2017plug,sun2016deep,adler2017learned,diamond2017unrolled,metzler2017learned,kim2016deeply,cascade_cnn_mrreocn_ismrm_2017,morteza2018neuralPGD,soltanayev2018training, metzler2018unsupervised,Bora_dimakis_2017} and references therein. The crux of unrolled schemes is to cast recovery as an image-to-image translation task mapping a low-quality image (e.g., found as linear estimate) to a high quality label image. It is essentially a cascade of alternating denoisers and data adjustment units with denoisers modeled via neural networks.

The denoiser networks are either allowed to be shared, leading to RNN architectures (see e.g.,~\cite{morteza2018neuralPGD, kim2016deeply}), or separately vary across iterations (see e.g.,~\cite{sun2016deep, diamond2017unrolled, cascade_cnn_mrreocn_ismrm_2017}). The latter seems to entail smaller networks that are easier to train, and they have been observed to achieve promising generalization performance for solving inverse problems. A noteworthy example of which includes the neural proximal gradient descent algorithm (NPGD) that models the proximal map using residual networks (ResNets) and after training using a recurrent neural network (RNN) it achieves state-of-the-art performance for compressed sensing tasks such as MR imaging. The recurrent inference machines in \cite{putzky2017recurrent} also leverages RNNs to learn the prior distribution parameters. For natural image superresoltuion tasks also cross-layer weight sharing achieves state-of-the-art quality~\cite{kim2016deeply}.

All in all, the generalization risk of unrolled neural networks for inverse problems has not been studied to date. This paper aims to extensively study this and reveals the influence of network architecture and train sample size.






\noindent\textbf{Contributions.}~
In order to study the prediction error of unrolled neural networks we leverage the Stein's Unbiased Risk Estimator (SURE)~\cite{stein1981estimation,donoho1995adapting} as a proxy for the generalization MSE. In statistical estimation, SURE comprises two terms, residual sum of squares (RSS) plus degrees of freedom (DOF), where RSS typically accounts for the prediction bias, while DOF captures the prediction uncertainity. We particularly focus on DOF, that is the trace of the end-to-end network Jacobian. Adopting a single layer residual unit with skip connection for the denoiser, the achievable DOF is analyzed for the denoising task. Under certain incoherence conditions on the weights of the trained network, DOF is derived in terms of the weighted path sparsity of the network activation as a useful surrogate to assess the generalization risk.

We conducted extensive empirical evaluations for denoising natural images, confirming the theoretically-predicted behavior of SURE. The adopted RNN architecture with weight sharing (WS) is compared with an alternative scheme where the weights are allowed to freely change across iterations, the so termed weight changing (WC). The comparisons reveal an interesting trade off. WS achieves higher DOF but lower RSS than WC. The overall SURE for WC is however smaller. The SURE gap between WS and WC schemes is shown to be significant for low sample sizes, but decreases as the sample size grows; eventually WS and WC agree as label scarcity abates. Further experiments for natural image deblurring show superior PSNR for WS  vs. WC. We also compared the filtering behavior of the learned proximals for WS and WC inspired by deep scattering networks~\cite{bruna2013invariant,anden2014deep}. For this purpose we analyzed the frequency spectrum of different iterations that show WS performs bandpass filtering, while WC alternates between low and bandpass filtering to denoise the images.

In summary, these findings rest on several novel contributions:

\begin{itemize}
    \item Theoretical analysis of generalization risk using SURE for recurrent unrolled networks
    
    
    \item Proved that DOF is well-approximated by the weighted \textit{path sparsity} under proper incoherence conditions 
    
    \item Extensive empirical evaluation of SURE for recurrent and non-recurrent networks for natural image denoisng and deblurring (compressed sensing MRI in supplementary materials). 
    
    \item Filtering interpretation of the trained weights in recurrent and non-recurrent unrolled networks.

    
\end{itemize}

The rest of this paper is organized as follows. Section 2 introduces the preliminaries on neural proximal algorithms and states the problem. The generalization risk is analyzed in Section 3. Empirical evaluations are then reported in Section 4, while Section 5 discusses the conclusions. 

\noindent\textit{Notations.}~In the paper $(\cdot)^{\dagger}$, $( \cdot)^{\mathsf{H}}$, $\|\cdot\|_2$, $\EE$, $\tr$, $\circ$, and $I_n$ refer to the matrix pseudo inverse, Hermitian, $\ell_2$-norm, statistical expectation, trace, composition operator, and $n \times n$ identity matrix. $\mathbbm{1}_{\{x\}}$ also denotes the indicator function that is unity for $x>0$, and zero otherwise.

\section{Preliminaries and Problem Statement}
\label{prelims}
\vspace{-2mm}
Consider the linear system

\vspace{-4mm}
\begin{equation}
y = \Phi x+ v  \label{eq:linear_system}
\end{equation}
\vspace{-5mm}

with $\Phi \in \mathbbm{C}^{m \times n}$ and $m \leq n$, where the Gaussian noise $v \sim \mathcal{N}(0,\sigma^2)$ captures the noise and unmodeled dynamics. Suppose the unknown image $x$ lies in a low-dimensional manifold. No information is known about the manifold besides the training samples $\cX:=\{x_i\}_{i=1}^N$ drawn from it, and the corresponding noisy observations $\mathcal{Y}:=\{y_i\}_{i=1}^N$. Given a new undersampled observation $y$, the goal is to quickly recover a plausible image $\hat{x}$ that is close to $x$.

The stated problem covers a wide range of image recovery tasks. For instance, for image denosing $\Phi=I$~\cite{dabov2007image,dong2013nonlocally,zhang2017beyond}, for image deblurring $\Phi$~\cite{zoran2011learning,venkatakrishnan2013plug} signifies the local convolution operator, for image superresolution~\cite{romano2017raisr,bruna2015super} $\Phi$ is the downsampling operator that averages out nonoverlapping image regions to arrive at a low resolution image, and for compressed sensing MRI~\cite{pualy_mri20017} $\Phi$ refers to the subsapmed Fourier operator.

\subsection{Neural proximal learning}
\vspace{-2mm}

%
In order to invert the linear system \eqref{eq:linear_system} a variation of the proximal algorithm advocated in \cite{morteza2018neuralPGD} is adopted. Given a pre-trained proximal operator $\cP_{\psi}$~\cite{parikh2014proximal} modeled via a neural network, the overall iterative procedure evolves according to the state-space equations
\[
\begin{array}{cc}
     & \hspace{-2.2cm}  {\rm \textbf{step 1.}} \quad\quad\quad s^{t+1} = g\big(x^{t};y \big) \\
     & \hspace{-2cm} {\rm \textbf{step 2.}} \quad\quad\quad x^{t+1} = \cP_{\psi}(s^{t+1})
\end{array}
\]
for a fixed number of iterations, i.e., $t=1,\ldots,T$. The first step invokes a linear operation that assures the state consistency with the measuremnts $y$. The second step executes the proximal mapping for denoising the image estimate. The recursion starts with the initial linear estimate ${x}^{0} = \Phi^{\mathsf{H}} y$ as the match filtered input $y$. For the first step, we can perform a first-order gradient step as in \cite{morteza2018neuralPGD}, or (preferably) a second-order least-squares step if computationally affordable. They are expressed as follows (for a learnable step size $\alpha$):
%
\begin{itemize}
\item Gradient step
\[ 
\hspace{-2.25cm}g(x^t;y):=
\alpha\Phi^{\mathsf{H}} y + (I -  \alpha \Phi^{\mathsf{H}}\Phi )x^{t}. \label{eq:gradient_step}
\]
\item Least-squares step 
\[
    g(x^t;y):= (\alpha\Phi \Phi^{\mathsf{H}} + (1-\alpha) I)^{-1} \left( \alpha \Phi^{\mathsf{H}} y +  (1-\alpha) x^t \right).   \label{eq:ls_step}
\]
\end{itemize}


\subsection{Proximal modeling with neural networks}
\vspace{-2mm}
A network with $K$ residual units (RU) is adopted to model the proximal map $\cP_{\psi}$. Adopting the ReLU activation $\sigma(x)= D(x) \cdot x$, where $D(x) = \mathbbm{1}_{x},$ the outer iteration $t$ (mapping $x^{t-1}$ to $x^
{t}$) can be decomposed as follows,
\begin{itemize}
    \item $h^t_0 = g(x_{t-1};y)$
    \item ${h}^t_{k+1} = {h}^t_{k} + W_k^{\mathsf{H}} \sigma (\bar{W}_k  {h}^t_{k} ), \quad k = 1,\ldots, K$
    \item $x^t = h^t_K.$
\end{itemize}

Neural proximal algorithm is recurrent in nature to mimic the fixed point iteration for traditional proximal algorithm~\cite{parikh2014proximal}. We thus shared weights $\{W_k\}_{k=1}^K$ for different outer iterations $t$.  When $\bar{W}_k = {W}_k$ we call the model symmetric residual unit, which can provide further regularization through weight sharing.

However, we could also learn different weights $\{W^t_k\}_{k=1}^K$ for different $t$, 
which changes the hidden layers at $t$-th iteration to
\begin{equation}
        {h}^t_{k+1} = {h}^t_{k} + W_k^{t,\mathsf{H}} \sigma (\bar{W}^t_k  {h}^t_{k} ), \quad k = 1,\ldots, K
\end{equation}
This scheme known as weight changing is used later in the numerical experiments as the benchmark for performance comparison.

\noindent\textbf{Pseudo linear representation.}~We adopt a pseudo-linear representation for the activation, where $D_k$ is a diagonal mask matrix with binary values for ReLU. Note, during inference, the mask $D_k$ is dependent on the input data examples, while $W_k$ is fixed. Accordingly, we can write $h_{k+1}^t=M_k^t h_{k}^t$, where $M_k^t = I + W_k^{\mathsf{H}} D_k^t \bar{W}_k$. The overall proximal map $M_t$ at $t$-th iteration then admits
\begin{equation}
x^{t+1} = \underbrace{M_t^K \ldots M_t^2 M_t^1}_{:=M_t} s^{t+1}. \label{eq:output_klayer_net}
\end{equation}
Apparently, the map $M_t$ is input data dependent due to nonlinear activation.

Unrolling the $T$ outer iterations of the proximal algorithm and the $K$ inner iterations of the $K$ RUs, the end-to-end recurrent map with input $y_i$ yields
\begin{equation}
   \hat{x}_i :=x^T_i:= (M_T \circ g) \circ \ldots \circ (M_1 \circ g)(\Phi^{\mathsf{H}}y_i). \label{eq:hat_x_i}
\end{equation}

\noindent\textbf{Training.}~One optimizes the network weights $\mathcal{W}:=\{W_k\}_{k=1}^K$ to fit $\hat{\mathcal{X}}:=\{\hat{x}_i\}$ to $\mathcal{X}:=\{x_{i}\}$ for the training population using pixel-wise empirical loss
\begin{equation}
    \mbox{minimize}_{\mathcal{W}} \quad \frac{1}{N} \sum_{i=1}^N 
\|\hat{x}_i - x_{i} \|_2^2.
\end{equation}
%


\section{Risk Analysis}
\label{sec:risk}
\vspace{-2mm}

%
In order to ease the analytical exposition for the generalization risk we focus on the denoising task ($\Phi=I$),
\begin{equation}
    y= x + v,   \quad\quad v \sim \cN(0,\sigma^2)  \label{eq:denoise_model}
\end{equation}
with $\|x\|_2=1$. The derivation presented here could be generalized to an arbitrary $\Phi$ with noise following an exponential distribution using ideas presented in \cite{eldar2009generalized}. Let $x^T=h_{\Theta}(y)$ denote the prediction obtained via neural proximal algorithm for the function $h_{\Theta}(\cdot)$ in \eqref{eq:hat_x_i} with a test sample $y$ as input argument. Assume $h$ is weakly differentiable. The Stein's unbiased risk estimator (SURE)~\cite{stein1981estimation,donoho1995adapting} for $h_{\Theta}(y)$ is then expressed as
\begin{equation}
    {\rm SURE}(y)=  - n \sigma^2 + \underbrace{  \| h_{\Theta}(y) - y \|_2^2}_{{\rm RSS}(y)} + 2\sigma^2 \underbrace{\nabla_y \cdot h_{\Theta}(y)}_{{\rm DOF}(y)},  \label{eq:sure_h}
\end{equation}
where $\nabla$ is the divergence operator. Note, SURE has two nice properties. First, it does not depend on the ground truth $x$. Second, it is an unbiased estimate of the test mean-squared-error (MSE), namely
\begin{equation}
    {\rm MSE} := \EE \left[ \|h_{\Theta}(y)-x\|_2^2 \right] = \EE \left[{\rm SURE}(y)\right].
\end{equation}

SURE in \eqref{eq:sure_h} comprises two main terms. The residual sum of squares (RSS) measures the error between the corrupted and denoised input, while the DOF measures the \textit{achievable} degrees of freedom for the denoiser.


\noindent\textbf{Lemma 1}~\textit{For the considered neural proximal algorithm with the end-to-end nonlinear map $J_T$, namely $x^T = J_T y$ suppose that $\|x^T\|_2=1$ (after normalization). It then holds that}
\[ 
\begin{array}{ll}
{\rm DOF} &= \EE  [\tr(J_T)]\\

    {\rm RSS} 
    &= n \sigma^2 + 2 (1 -  \EE[ \tr(y^{\mathsf{H}} J_T y)])
\end{array}
\]


%
A natural question then pertains to the behavior of DOF and RSS terms as well as the overall SURE. In particular, we are interested in DOF that is known in statistical estimation as a measure of the predictor variability and uncertainty~\cite{stein1981estimation}. It has been adopted as a notion of optimism or prediction error in~\cite{efron2004estimation}. Parameter tuning using SURE has also been advocated in~\cite{donoho1994threshold, donoho1995adapting}. SURE has also been used recently for unsupervised training of deep inverse maps~\cite{metzler2018unsupervised,soltanayev2018training}. DOF captures the neural network capacity (free parameters), the number of training samples, and the training algorithm. A rigorous analysis of DOF for the trained neural PGD algorithm is the subject of next section.




\subsection{Degrees of freedom}
\label{sure}
\vspace{-2mm}

%
To facilitate the SURE analysis, we consider the proximal map with a single RU. We assume that RU is symmetric, meaning that the deconvolution operation is simply the convolution transposed with a negative scaling. The corresponding proximals then simply admit $M_t =  I - W^{\mathsf{H}} D_t W$ at $t$-th iteration~(see \eqref{eq:output_klayer_net}). In addition, assume that for the gradient step $\alpha=0$, meaning that $g(x^t;y)=x^t$. One can alternatively encourage data consistency by imposing data fidelity during the training. Accordingly, from \eqref{eq:hat_x_i} it is easy to derive the end-to-end map relating the iteration outputs $x^{t+1}$ to the initial estimate $x^0=y$ as
\begin{align}
        {x}_{t+1} & = \underbrace{(I - W
^{\mathsf{H}} D_{t+1} W)  \ldots (I - W
^{\mathsf{H}} D_{1} W)}_{:=J_{t+1}} y         
\end{align}
%
 One can then expand $J_T$ as a linear combination of $J_{\cI}$'s where there are in total $2^T$ different index choices $\cI$ that can be associated with an index array $(i_1, \ldots, i_{j})$. Accordingly,
%
\begin{align}
J_T &= I + \sum_{\cI} (-1)^{|\cI|} J_{\cI},\\
J_{\cI} &= W^{\mathsf{H}} D_{i_j} W W^{\mathsf{H}} \ldots W W^{\mathsf{H}} D_{i_{1}} W, \label{eq:linear trace}
\end{align}

\subsubsection{Linear networks}
\label{subsubsec:linear_net}
To ease the exposition we begin with a linear network where the proximal map $\cP_{\Psi}(x)= (I-W^{\mathsf{H}}W)x$ is repeated for infinitely many iterations, namely $T \rightarrow \infty$. As a result of end-to-end training, the overall mapping is bounded and admits
\begin{align}
    \lim_{T \rightarrow \infty} J_T y = \underbrace{(I-W W^{\dagger})}_{:=J_{\infty}} y
\end{align}
where $\cP_W := W W^{\dagger}$ is orthogonal projection onto the range space of $W$.



The overall network mapping $J_{\infty}$ resembles singular value thresholding for denoising~\cite{donoho2014minimax}. The expcted DOF then admits a closed-form expression as stated next (proofs are included in the supplementary materials).


\noindent\textbf{Lemma 2.}~\textit{For the unrolled network with  single-layer linear residual units with $T \rightarrow \infty$, let $\{\sigma_i\}_{i=1}^n$ denote the singular values of the sample correlation matrix $C_x:=\frac{1}{N} \sum_{i=1}^N x_i x_i^{\mathsf{H}}$, and suppose $\hat{\sigma}^2 \approx \frac{1}{N}\sum_{i=1}^N v_i v_i^{\mathsf{H}}$. Then, DOF admits}
\begin{align*}
    &{\rm DOF} \approx n - \sum_{i=1}^{\min{\{\ell,n}\}} \mathbbm{I}_{\{\sigma_i - \hat{\sigma}^2  \}}
\end{align*}
%





\subsubsection{Non-linear networks}
\vspace{-2mm}

In the nonlinear setting to gain some intuition first we focus on the simpler case where the network entails infinitely many iterations. The end-to-end mapping is then expressed as
%

\begin{align}
  J_{T} y =    \prod_{\tau=0}^{T} (I- W^{\mathsf{H}} D_{T-\tau} W) y
\end{align}
for the mask sequence $\{D_t\}_{t=1}^{\infty}$. Assuming that the system returns a unique output, the mask sequence converges to the mask $\bar{D}$ of the latent code elements for the input $y$. Let the set $\cS$ include the support set for $\bar{D}$. Using similar arguments as the linear case it can be shown that the end-to-end mapping becomes a projection operator as follows
\begin{align}
    J_{\infty} y = \big(I-W_{\cS} {W_{\cS}}^{\dagger}\big)y
\end{align}
%



\noindent\textbf{Lemma 3.}~\textit{For an unrolled network with single-layer nonlinear residual units and $T \rightarrow \infty$ if the sequence of activation masks converge to $\bar{D}$ (for each input $y$), the DOF admits}
\begin{align}
      &{\rm DOF}= n - \EE[\tr(\bar{D})]
\end{align}
Apparently, the sparsity level of the last layers determines the DOF.

In practice, we are interested in understanding the behavior of unrolled neural networks with a finite number of iterations. Let $\rho_t=\EE[\tr(D_t)]$ be the expected sparsity level at $t$-th iteration. Define also the incoherence of the matrix $W$ as the largest off-diagonal inner product
\begin{align}
      \mu_W := \sup_{i\neq j}\big|[WW^{\mathsf{H}}]_{ij}\big|
\end{align}

For the trained network each noisy input activates neurons at certain hidden layers. Accordingly, one can imagine a connectivity graph through which the input pixels would traverse different paths to reach and form the output. As discussed earlier in \eqref{eq:linear trace} due to skip connections there are a total of $2^T$ possible paths. Introduce the diagonal matrix $B$ with the $i$-th diagonal element $[WW^{\mathsf{H}}]_{ii} = \| W_i\|_2^2$, and the cascade of the activation masks $D_{\cI} := D_{i_j} \ldots D_{i_1}$. Define then the weighted \textit{path sparsity}
\[ 
\begin{array}{ll}
p_{\cI} = \EE [\tr(D_{\cI}B^{|\cI|})]. 
\end{array}
\]
%


The following Lemma then bounds the deviation of the individual terms (associated with different paths) in the Jacobian expansion \eqref{eq:linear trace} from the neural network with orthogonal weight matrices.

\noindent\textbf{Lemma 4}.~\textit{For the trained network with the expected sparsity levels $\rho_i := \EE[\tr(D_i)]$ at $i$-th iteration, for the index subset $(i_1, \ldots, i_j)$, it holds that}

 \[
 \begin{array}{ll}
 \Big| \EE [ \tr(W^{\mathsf{H}} D_{i_j} W W^{\mathsf{H}} \ldots W W^{\mathsf{H}} D_{i_{1}} W) ]
-     p_{\cI} \Big| \le  \prod_{l=1}^j[\sqrt{s_{i_l}} (s_{i_l}-1)\mu_{W} ]   
\end{array}
\]

Combining Lemma 4 with \eqref{eq:linear trace}, the main result is established as follows.

\noindent\textbf{Theorem 1}.~\textit{For an unrolled network with a cascade of $T$ recurrent single-layer residual units let $\{\rho_t=\EE[\tr(D_t)]\}_{t=1}^T$. If the network weights satisfy $ \mu_W \le \epsilon \max_t \rho_t^{-3/2}$ for a constant $\epsilon<1$, the DOF is bounded as}
\[
\begin{array}{ll}
\Big| \EE[\tr(J_{T})] - n - \sum_{\cI} (-1)^{|\cI|} p_{\cI}  \Big| \le  (1+  \epsilon)^T -1-\epsilon T. 
\end{array}
\]

%
Accordingly, one can adopt $n+\sum_{\cI} (-1)^{|\cI|} p_{\cI}$ as a surrogate for DOF of the trained network.

\section{Empirical Evaluations}
\label{evals}
\vspace{-2mm}

Extensive experiments were performed to assess our findings for natural image denoising and deblurring. In particular, we aimed to address the following important questions:

\noindent\textit{Q1.~How would the RSS, DOF, and SURE behave empirically for WS and WC schemes?}

\noindent\textit{Q2.~How would MSE/PSNR behave as a function of the train sample size for WS and WC schemes?}

\noindent\textit{Q3.~What is the filtering interpreation of the learned denoisers for WS and WC schemes?}

\subsection{Network Architecture and Training}
\label{net_arch}
\vspace{-2mm}

To address the above questions, we adopted a ResNet with $2$ residual blocks (RBs) where each RB consists of two convolutional layers with $3 \times 3$ kernels and a fixed number ($128$) of feature maps, that were followed by batch normalization (BN) and ReLU activation. ResNet is used in the feature domain, and thus we add a convolutional layer with $3 \times 3$ kernels that lift up the image from previous iterations to $128$ feature maps. Similarly, ResNet is followed by a convolutional layer with $1 \times 1$ kernels that lifts off the feature maps to create the next estimate. We used the Adam optimizer \cite{kingma2014adam} with the momentum parameter $0.9$ and initial learning rate varying across the experiments.
Training was performed with TensorFlow and PyTorch interface on NVIDIA Titan X Pascal GPUs with 12GB RAM. PSNR (dB) is used as the figure of merit that is simply related to MSE as $PSNR=-10\log_{10}(MSE)$ since the images are normalized.

\subsection{Denoising}
\label{denoise}
\vspace{-2mm}

This section addresses Q1 and Q2 for natural image denoising task, where $\Phi = I$.

\noindent\textbf{Dataset.}~$400$ natural images of size $481 \times 321$ were selected from the Berkeley image segmentation dataset (a.k.a BSD68) \cite{martin2001database}. Patches of size $40 \times 40$ were extracted as labels, resulting in $230,400$ training samples. $68$ full images were chosen for test data.

The ResNet architecture described before was adopted with $K=2$ RBs and $T=4$ iterations. It is trained for $50$ epochs with minibatch size $256$. The initial learning rate was annealed by a factor of $10$ at the $40$-th epoch. We run experiments independently (with random initialization) for several initial learning rates, namely 0.0075, 0.005, 0.0025, 0.001, 0.00075, 0.0005, 0.00025, 0.0001, 0.000075, 0.00005, and pick the one leading to the best PSNR on test data during the last epoch. The aforementioned experiments were repeated for various noise levels $\sigma \in \{15,25,50,100\}$. Moreover, the experiments were repeated with and without weight sharing.

\begin{figure*}[t]
    \centering
    \begin{tabular}{c c c c }
        \centering
        \hspace{-3mm}\includegraphics[width=0.25\textwidth]{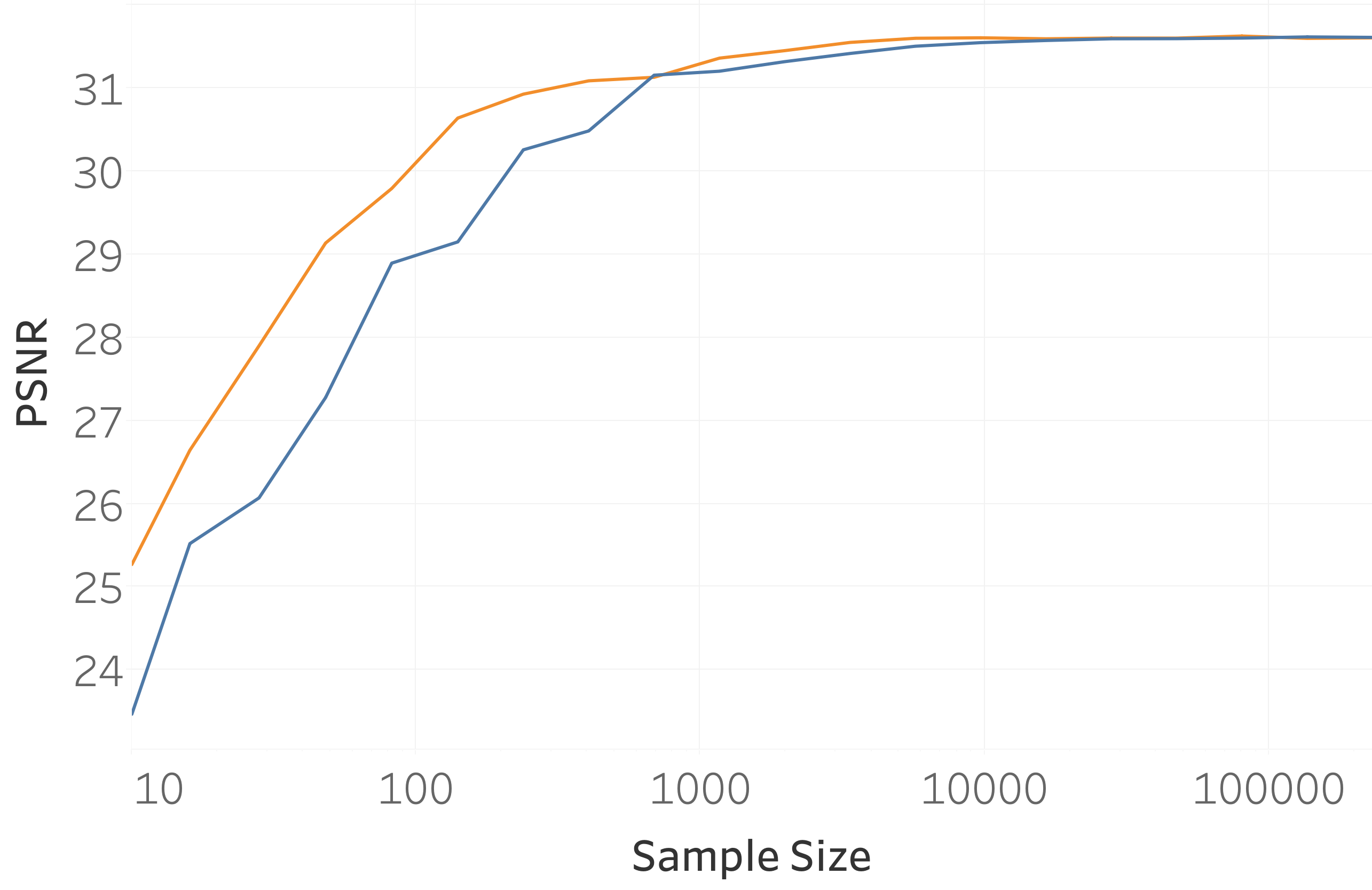}  &  

        \hspace{-3mm}\includegraphics[width=0.25\textwidth]{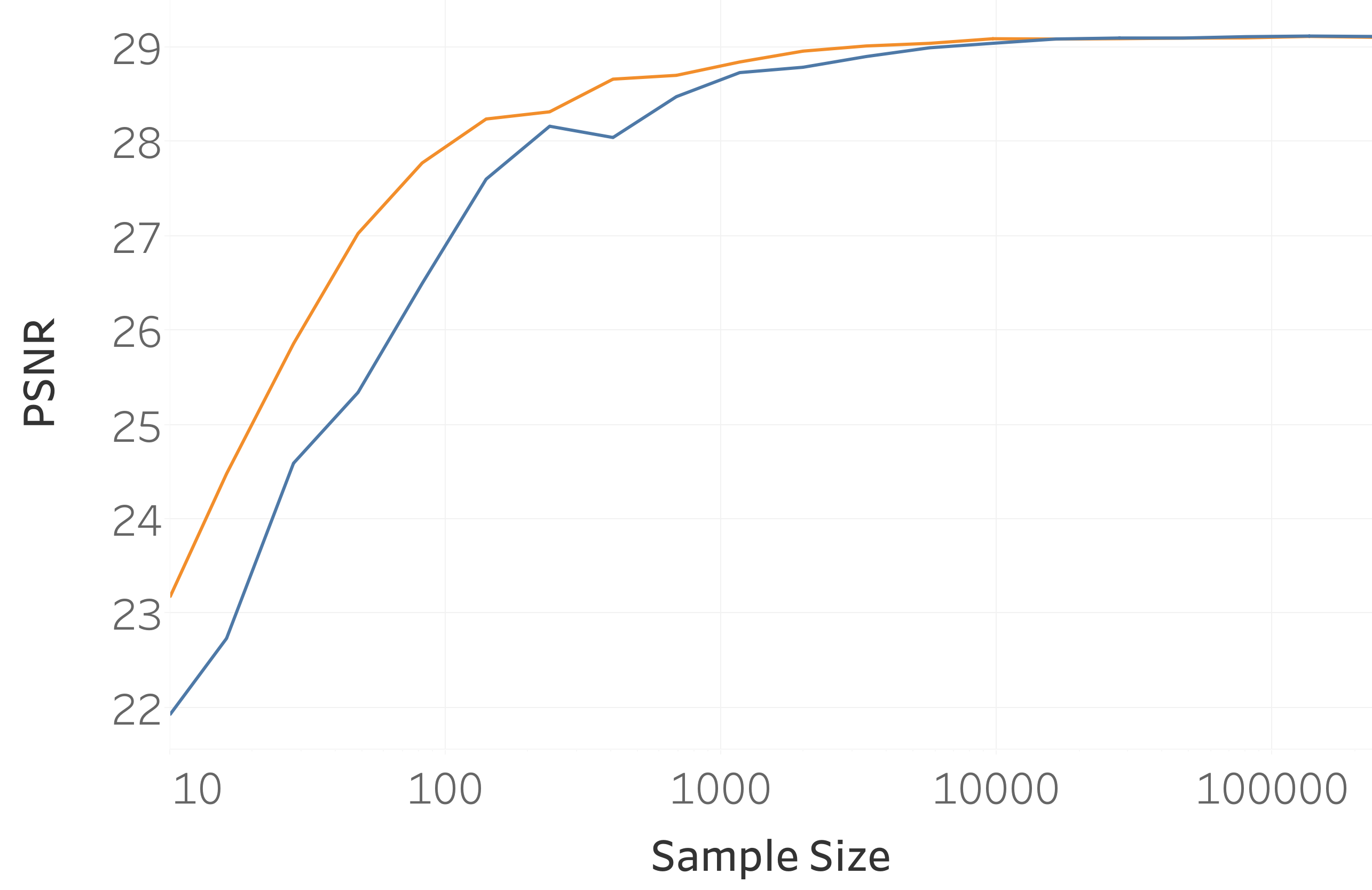}  &

        \hspace{-3mm}\includegraphics[width=0.25\textwidth]{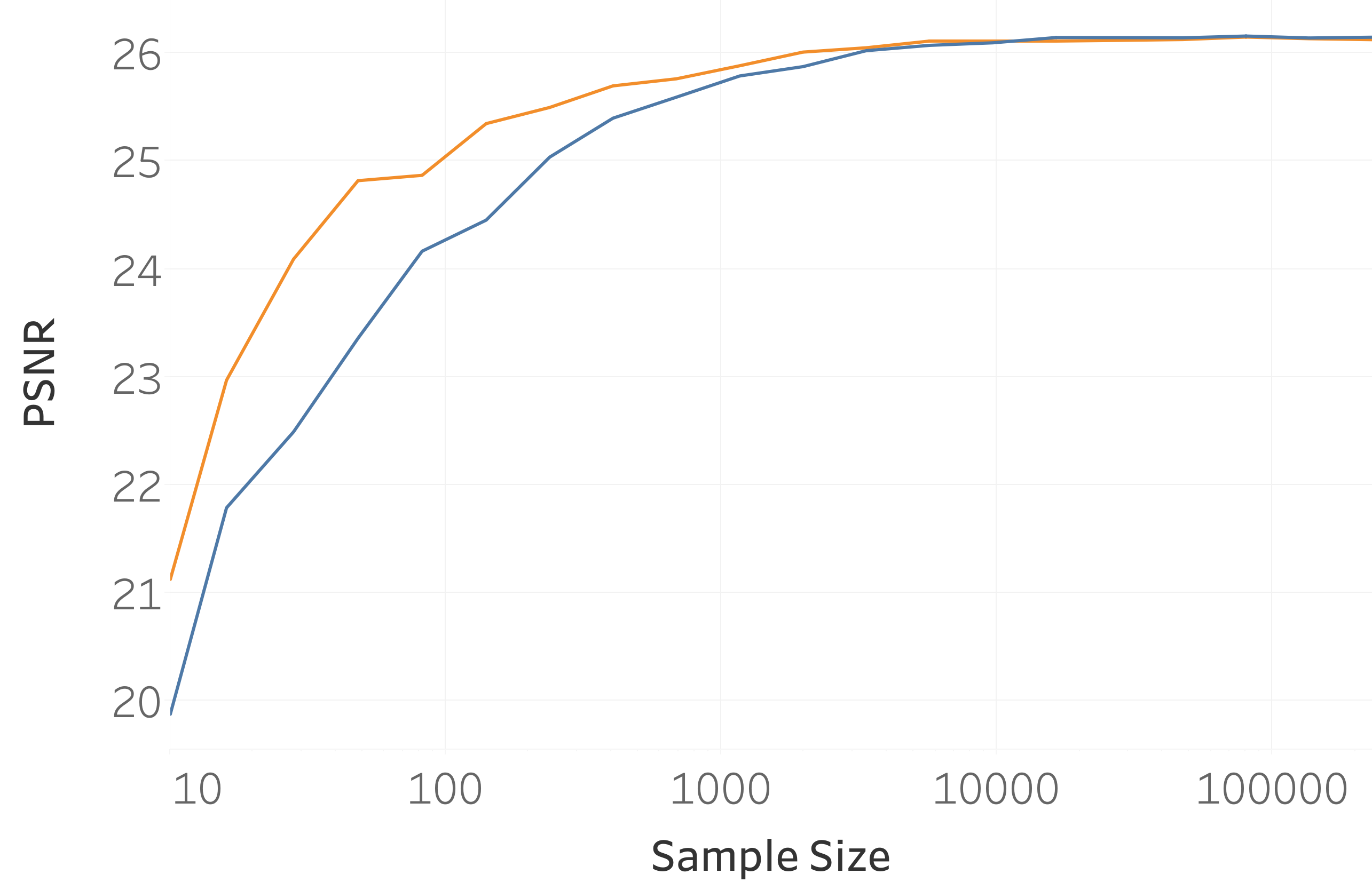}  &

        \hspace{-3mm}\includegraphics[width=0.25\textwidth]{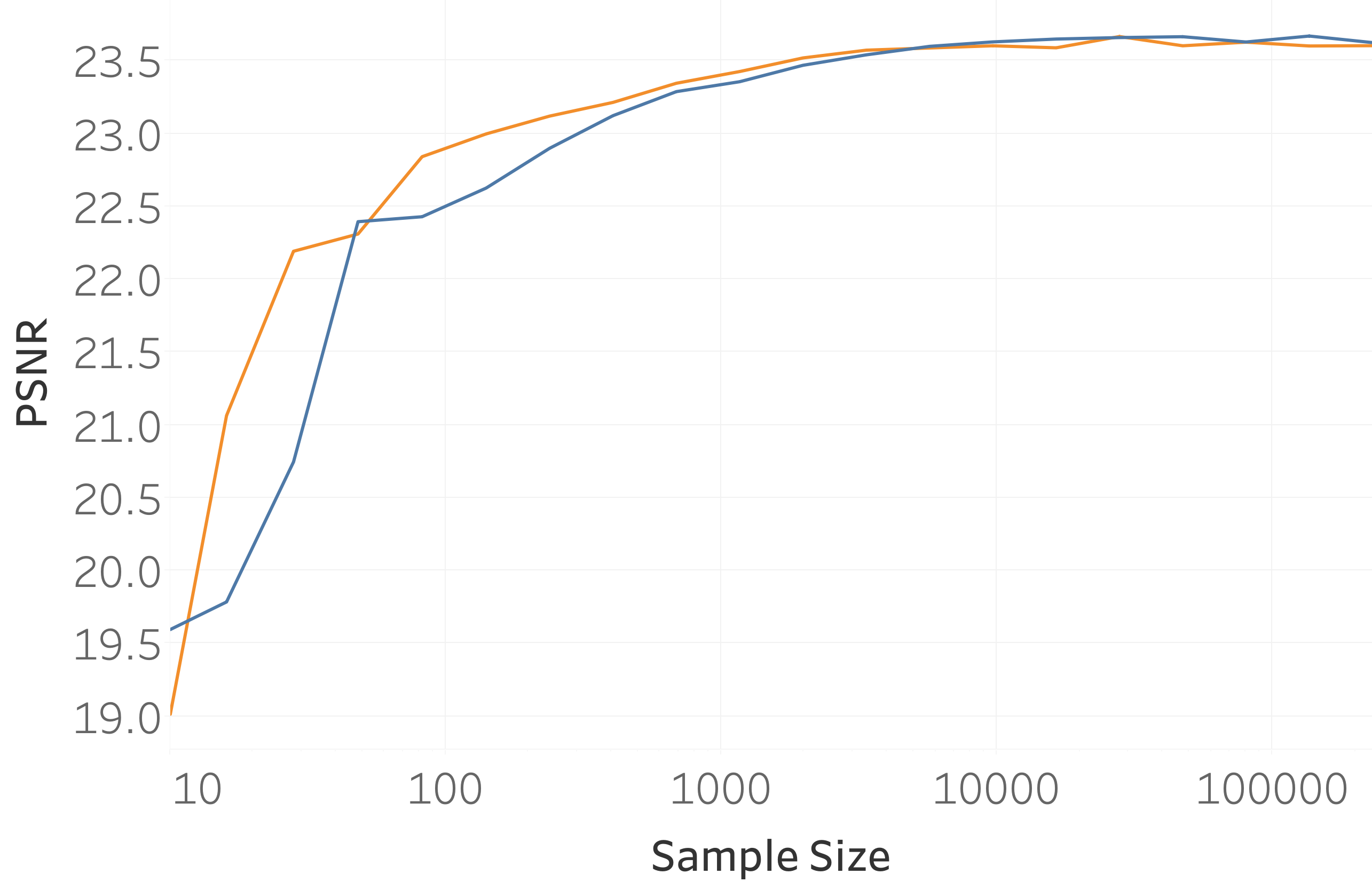}  \\
        
        \textbf{Denosing:\quad}($\sigma=15$)  &  ($\sigma=25$)  &   ($\sigma=50$)  &   ($\sigma=100$)  
        
        
    \end{tabular}
    
    \vspace{10mm}
    
    \begin{tabular}{c c}
        \includegraphics[width=0.25\textwidth]{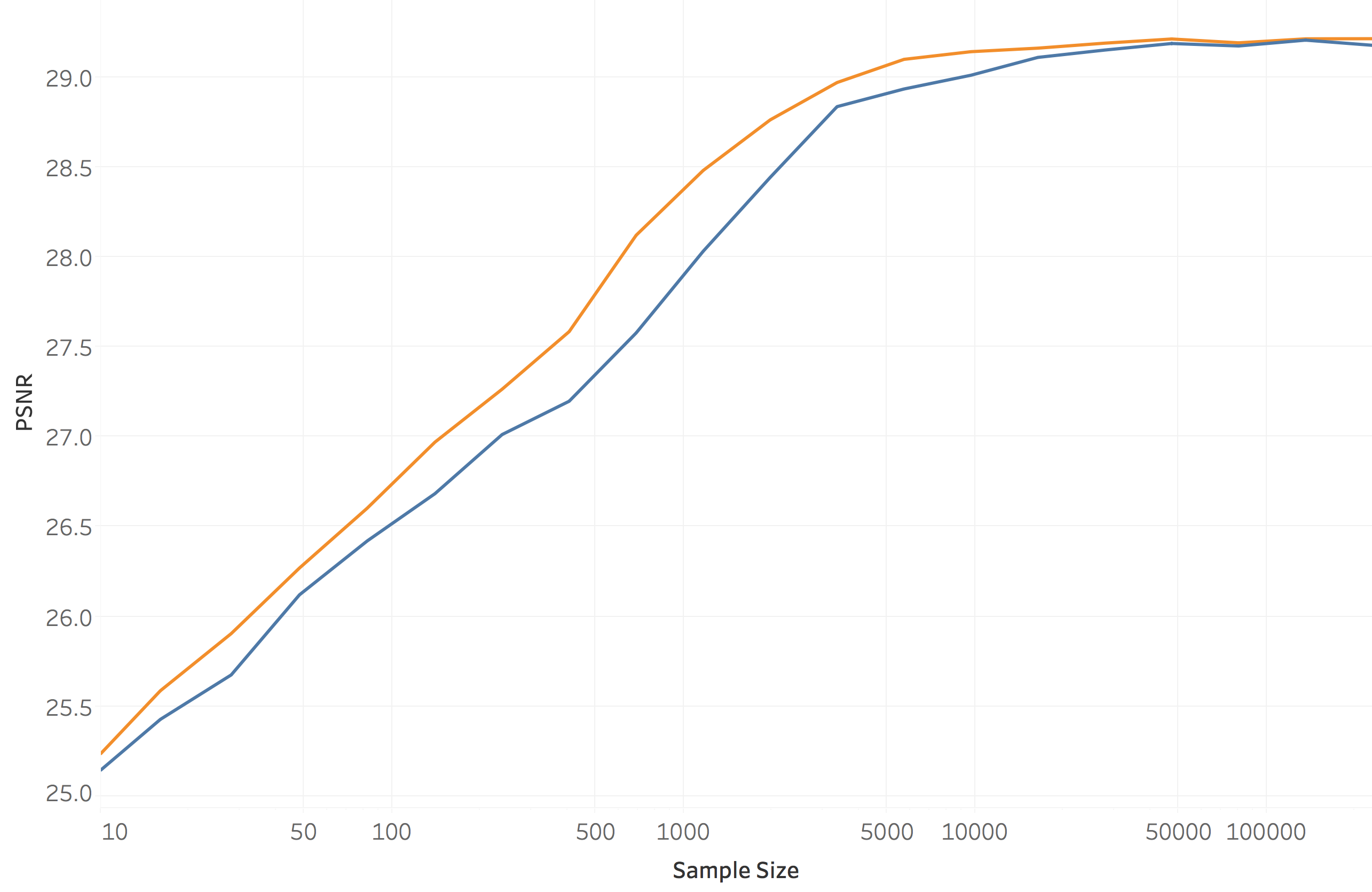}
        &
     \hspace{-0mm}
        \includegraphics[width=0.25\textwidth]{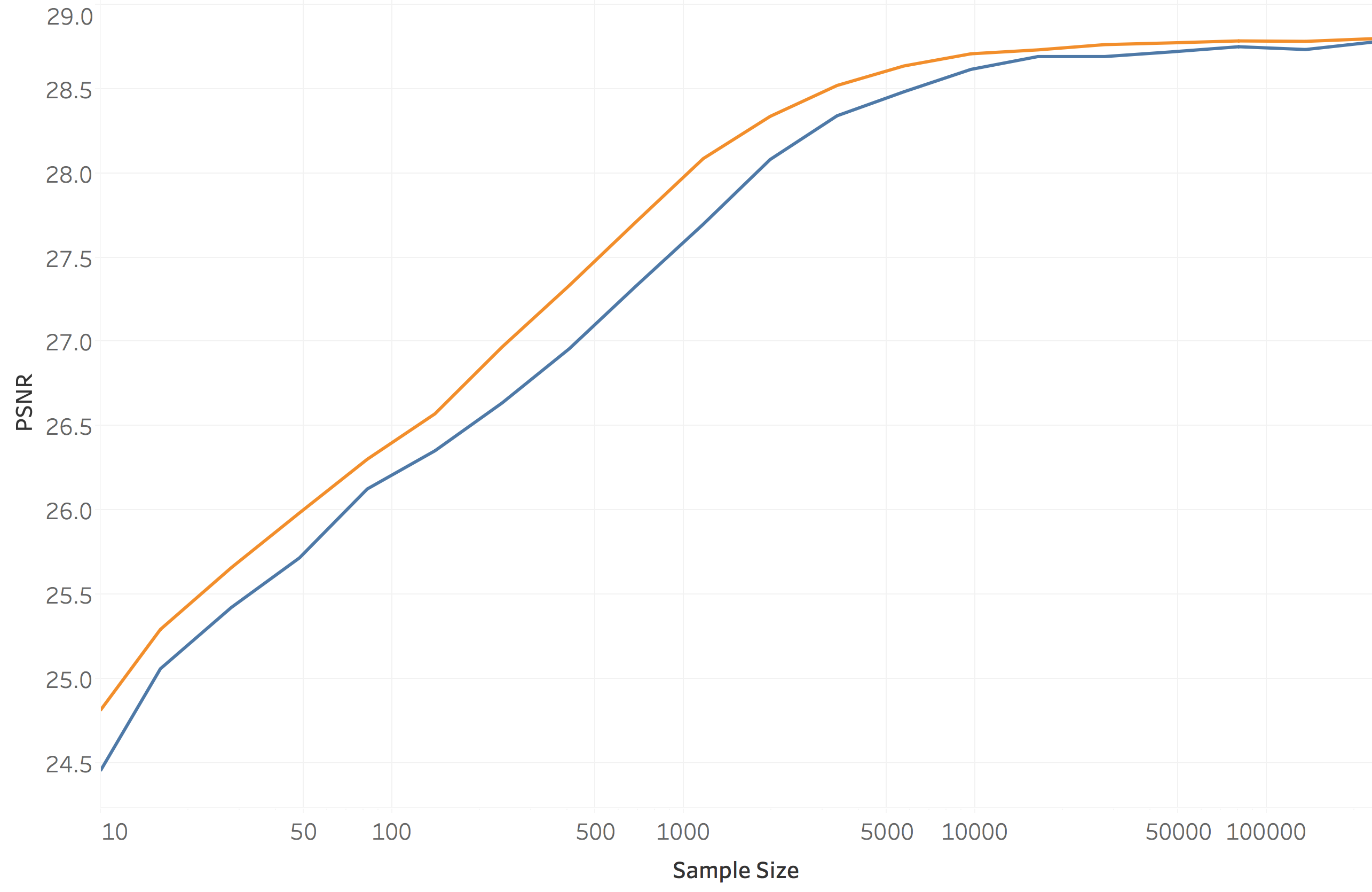} \\
        \textbf{Deblurring:\quad}( $\sigma=\sqrt{2}$)    &   ( $\sigma=2$)
\end{tabular}
    \caption{\textit{Effects of weight sharing and sample size on the image denoising and deblurring performance .} The upper row corresponds to denoising and the lower row corresponds to image deblurring. Different columns correspond to different noise levels. Each panel depicts the test PSNR as function of train sample size. The orange line corresponds to WS and the blue line to WC. 
    } 
    \label{fig:denoising}
\end{figure*}

We assess SURE, DOF, and RSS  with sample sizes within the range $[10,230400]$ (logarithmically spaced). It is first observed that the SURE estimate is in perfect agreement with the test MSE (or PSNR) when having the true labels available for validation purposes. We thus plot the PSNR evolution in Fig.~\ref{fig:denoising} as the train sample size grows (orange line for WS and the blue line for WC). For all noise levels, we observe a consistent benefit for WS in sample sizes less than $1$K. Interestingly, after $1$K they coincide and no benefit is observed for WC even for very large sample sizes in the order of $10^5$. Note also that the non-smooth behavior of the curve is mainly attributed to Adam optimizer that may not necessarily converge to the globally optimum network weights.

\begin{figure*}
    \centering
    \begin{tabular}{c c c c}
        \centering
        \includegraphics[width=0.22\textwidth]{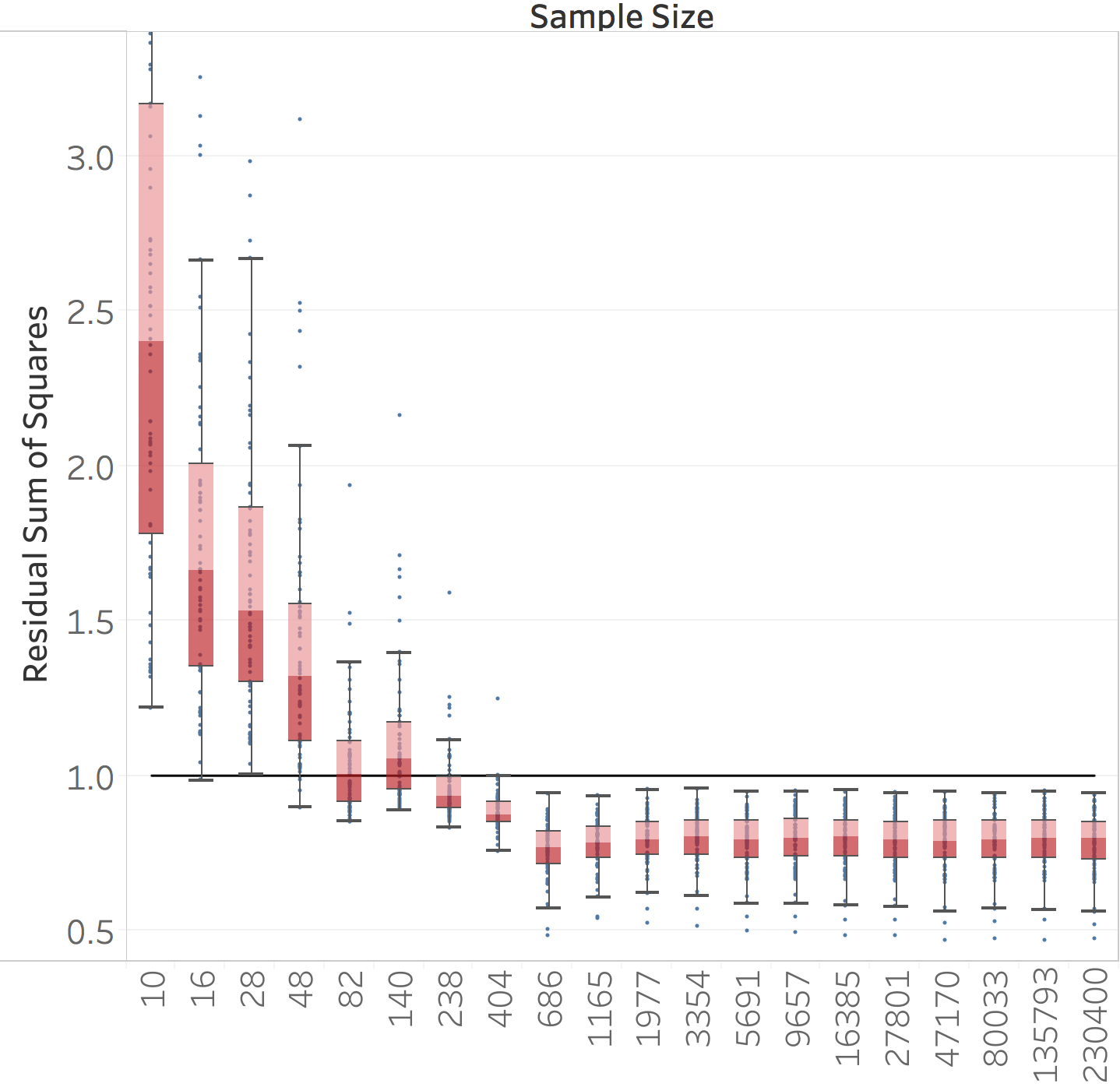}  &
        
        \includegraphics[width=0.22\textwidth]{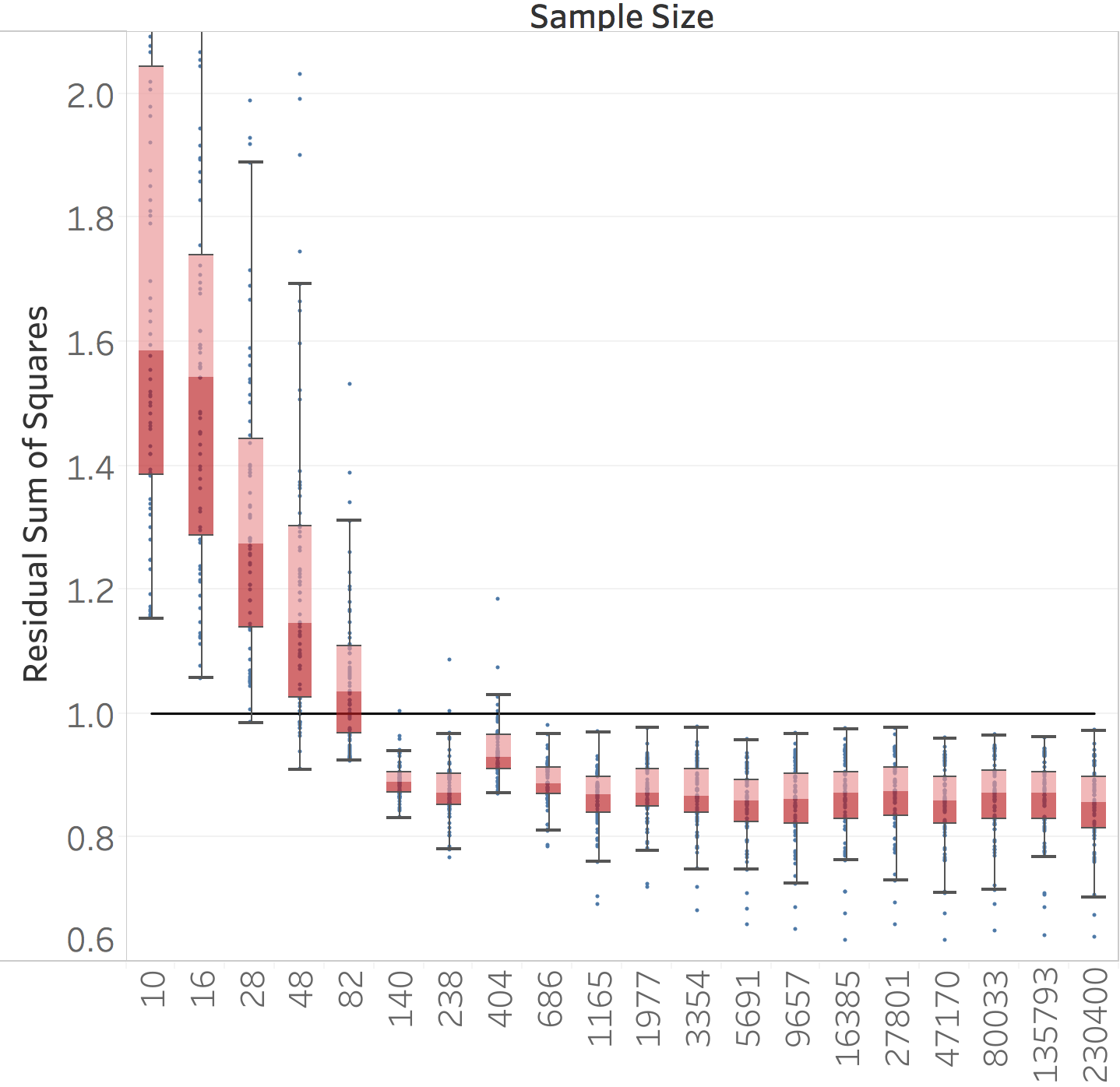}  &
        
        \includegraphics[width=0.22\textwidth]{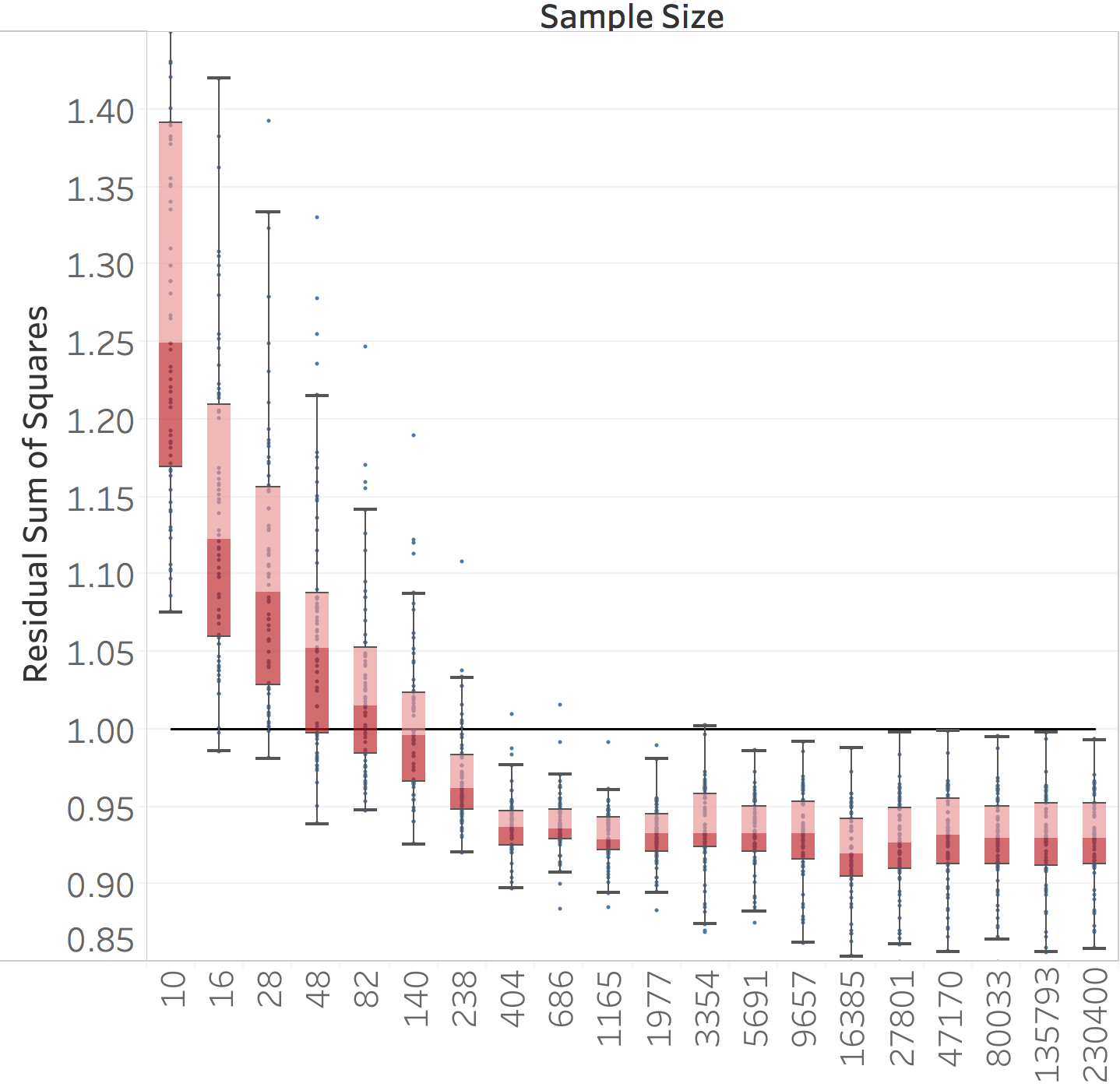}  &
        
        \includegraphics[width=0.22\textwidth]{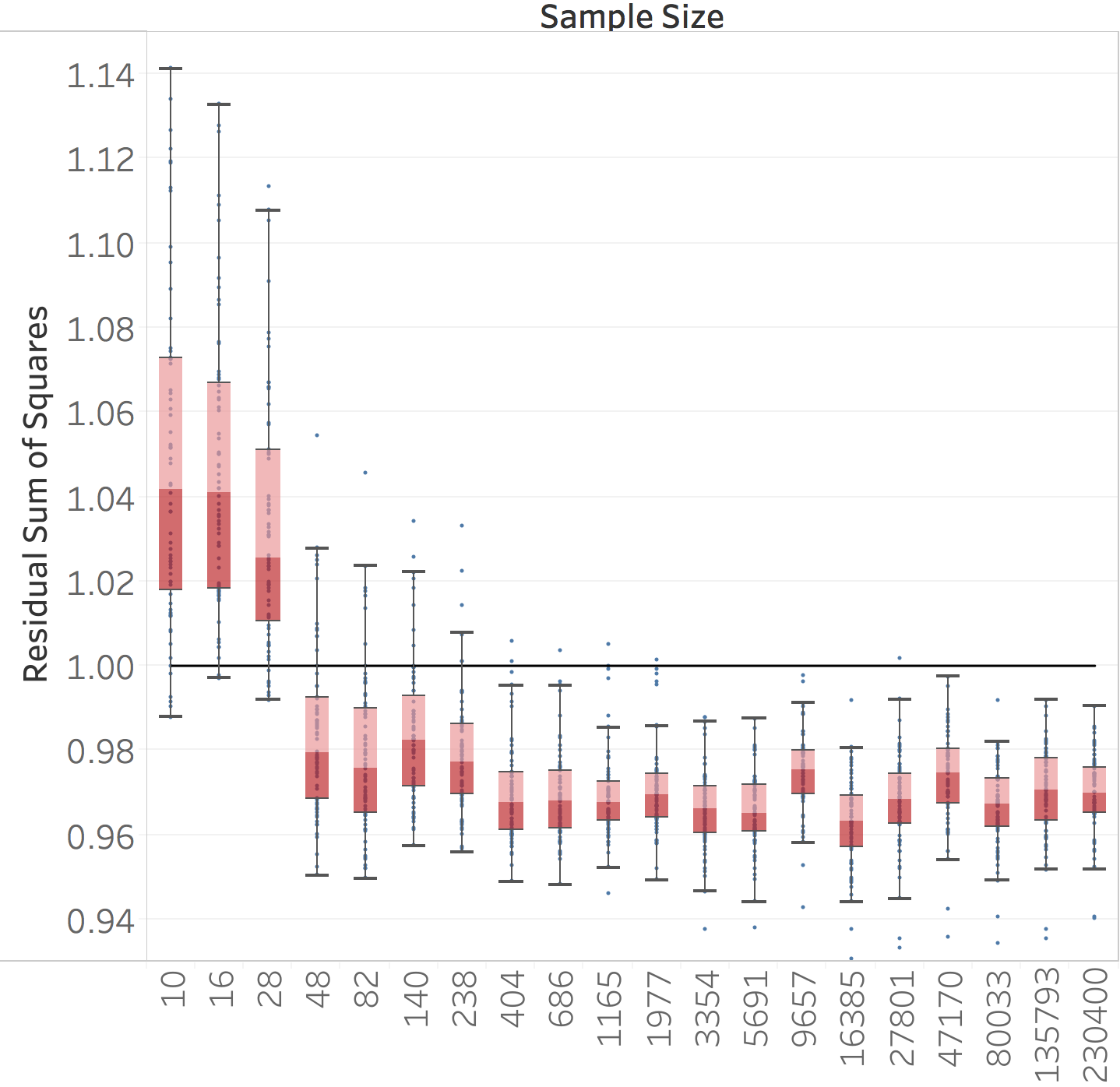}  \\

        \includegraphics[width=0.22\textwidth]{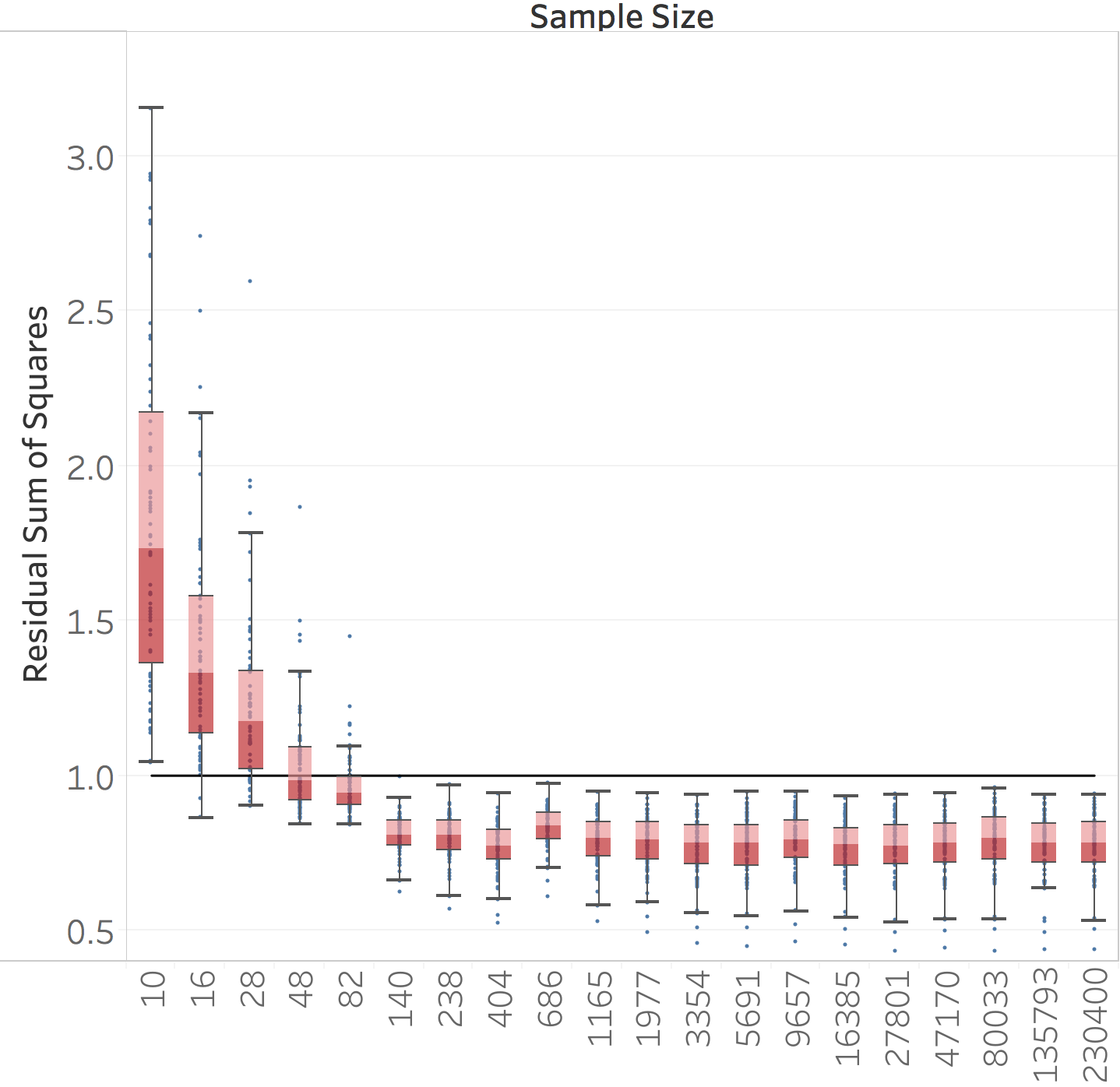} &

        \includegraphics[width=0.22\textwidth]{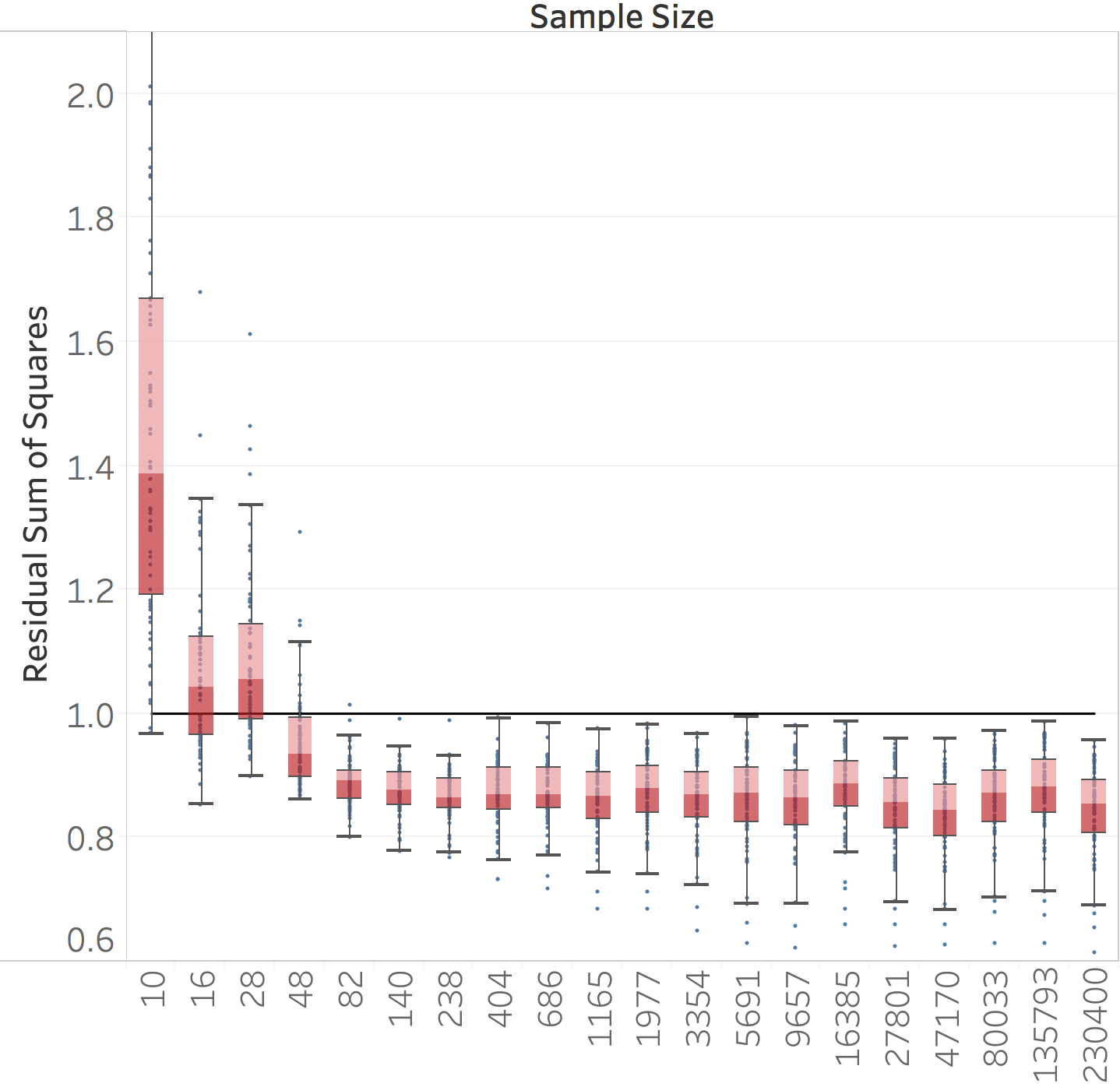}  &

        \includegraphics[width=0.22\textwidth]{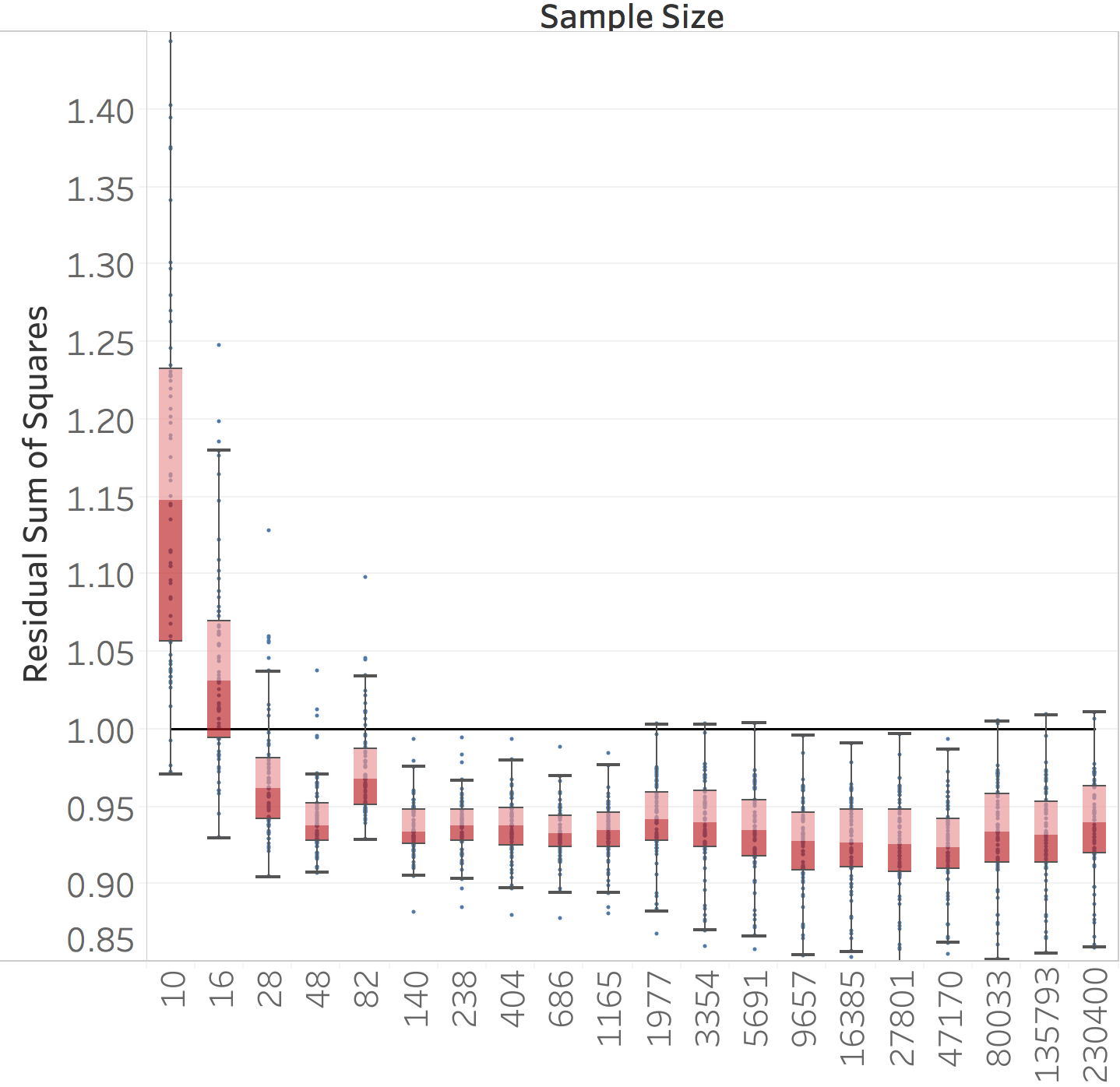} &

        \includegraphics[width=0.22\textwidth]{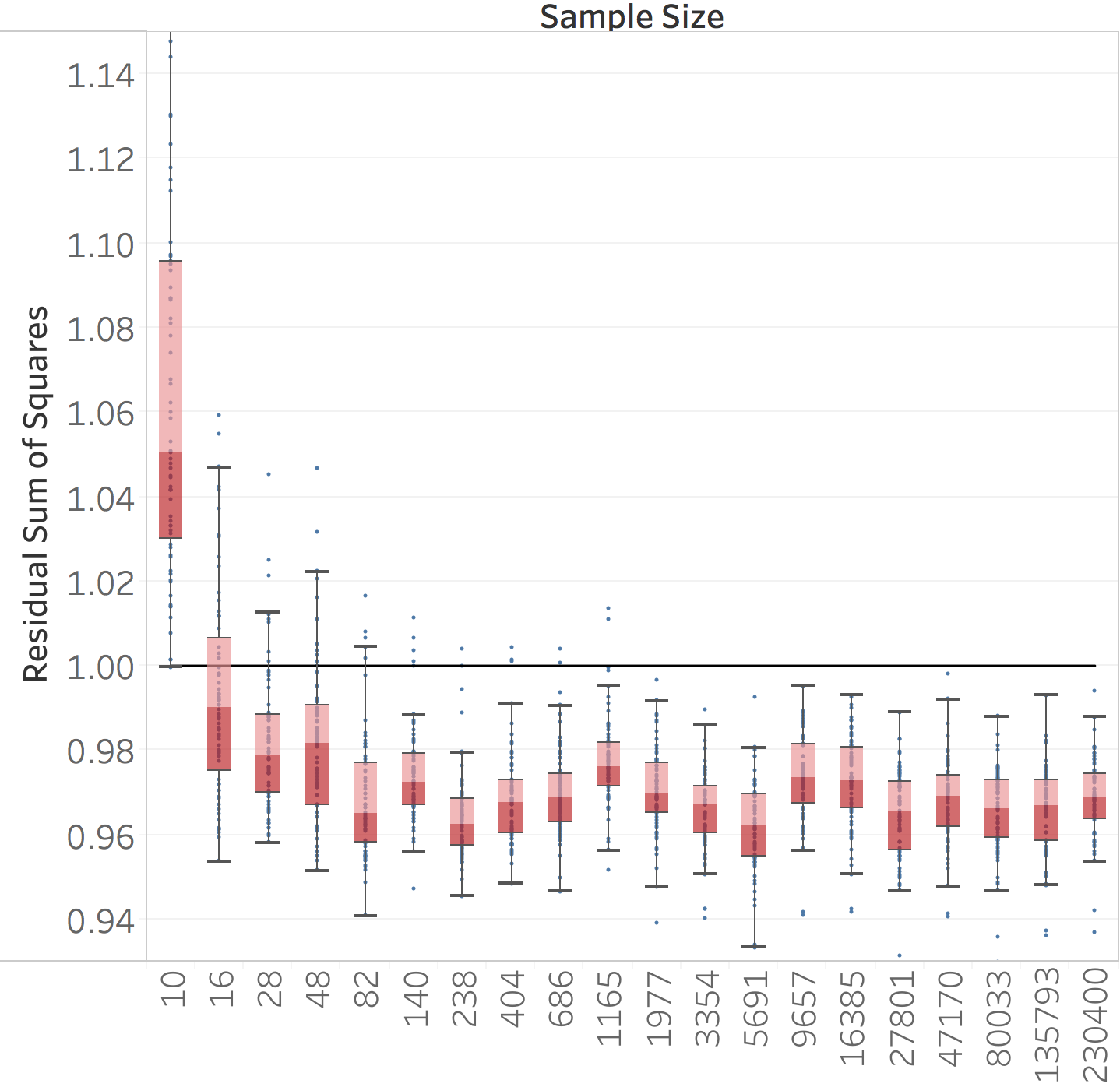} \\  
        
        \textbf{RSS:} ($\sigma=15$) & ($\sigma=25$) & ($\sigma=50$) & ($\sigma=100$)

    \end{tabular}
    
            \vspace{0mm}

     \begin{tabular}{c c c c}
        \centering
        \includegraphics[width=0.225\textwidth]{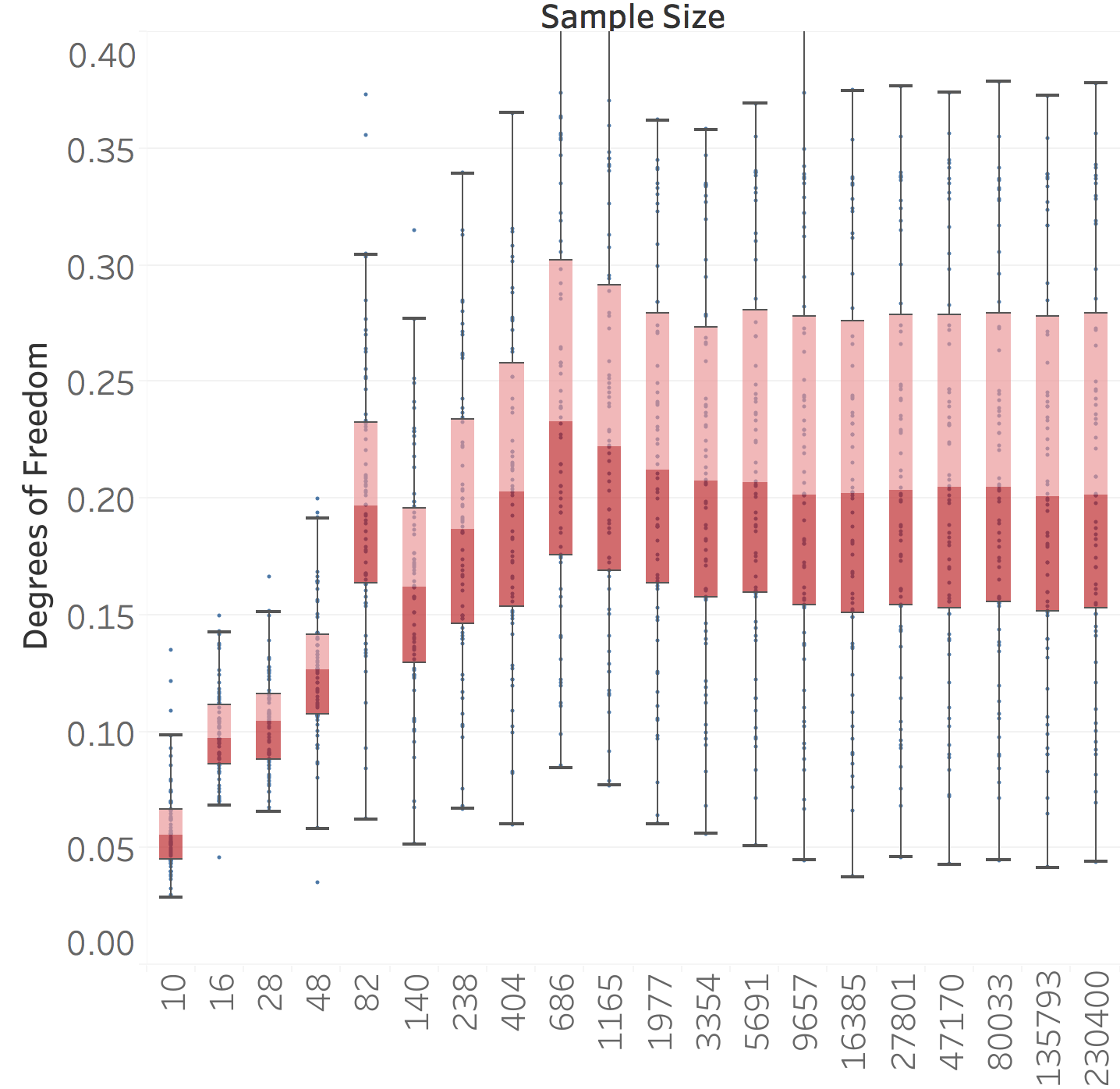} &

        \includegraphics[width=0.225\textwidth]{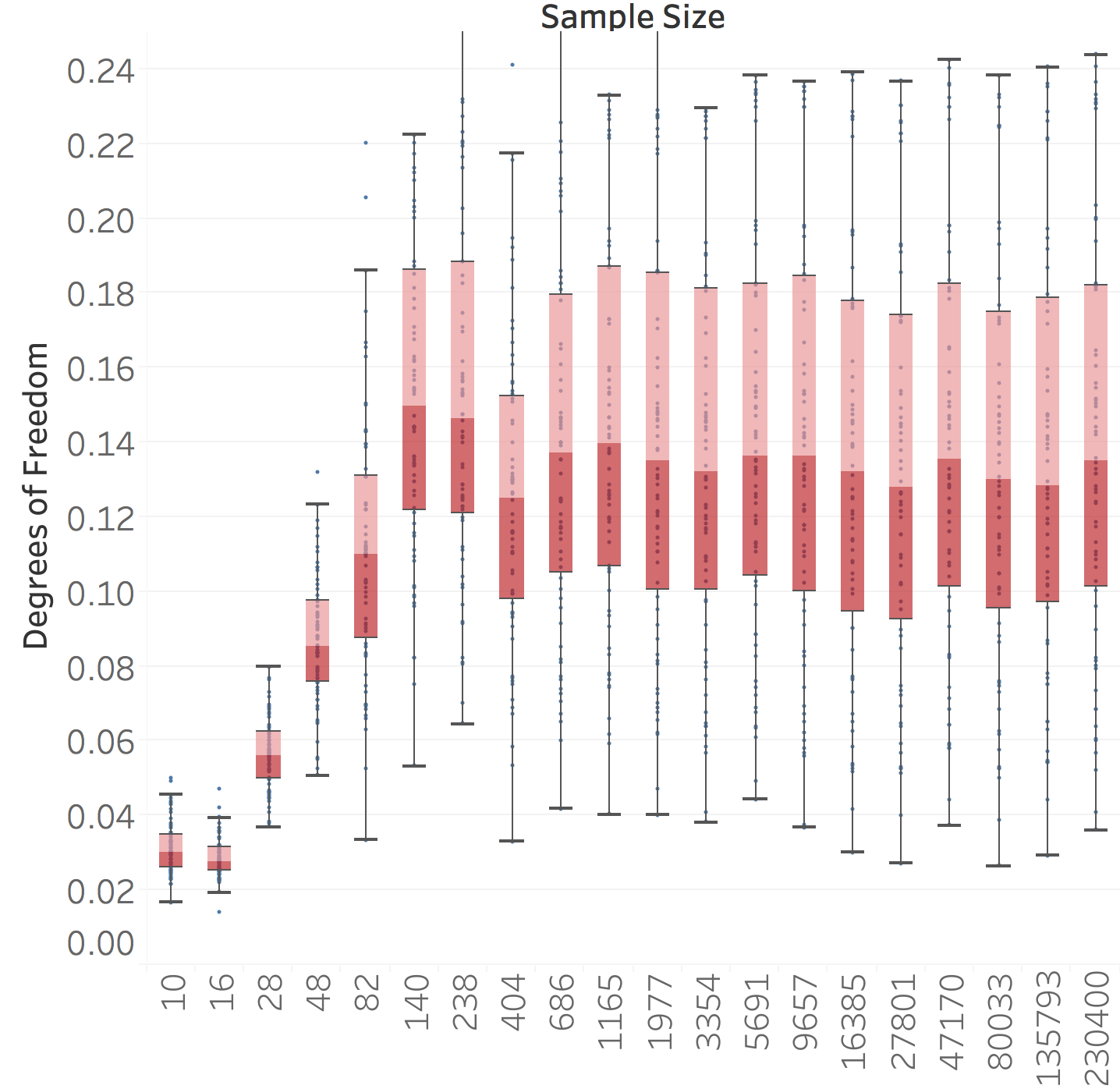} &

        \includegraphics[width=0.225\textwidth]{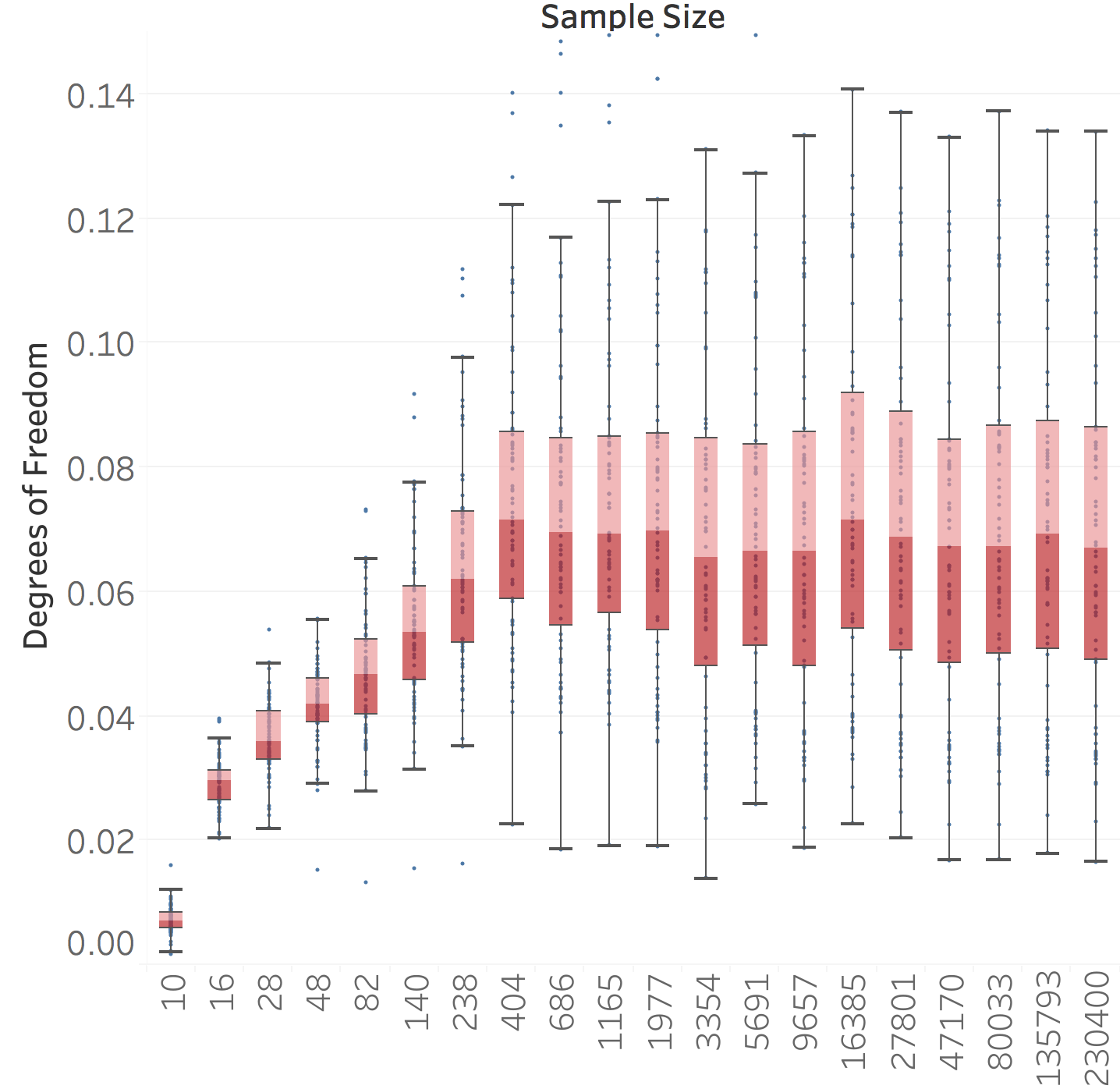}  &

        \includegraphics[width=0.225\textwidth]{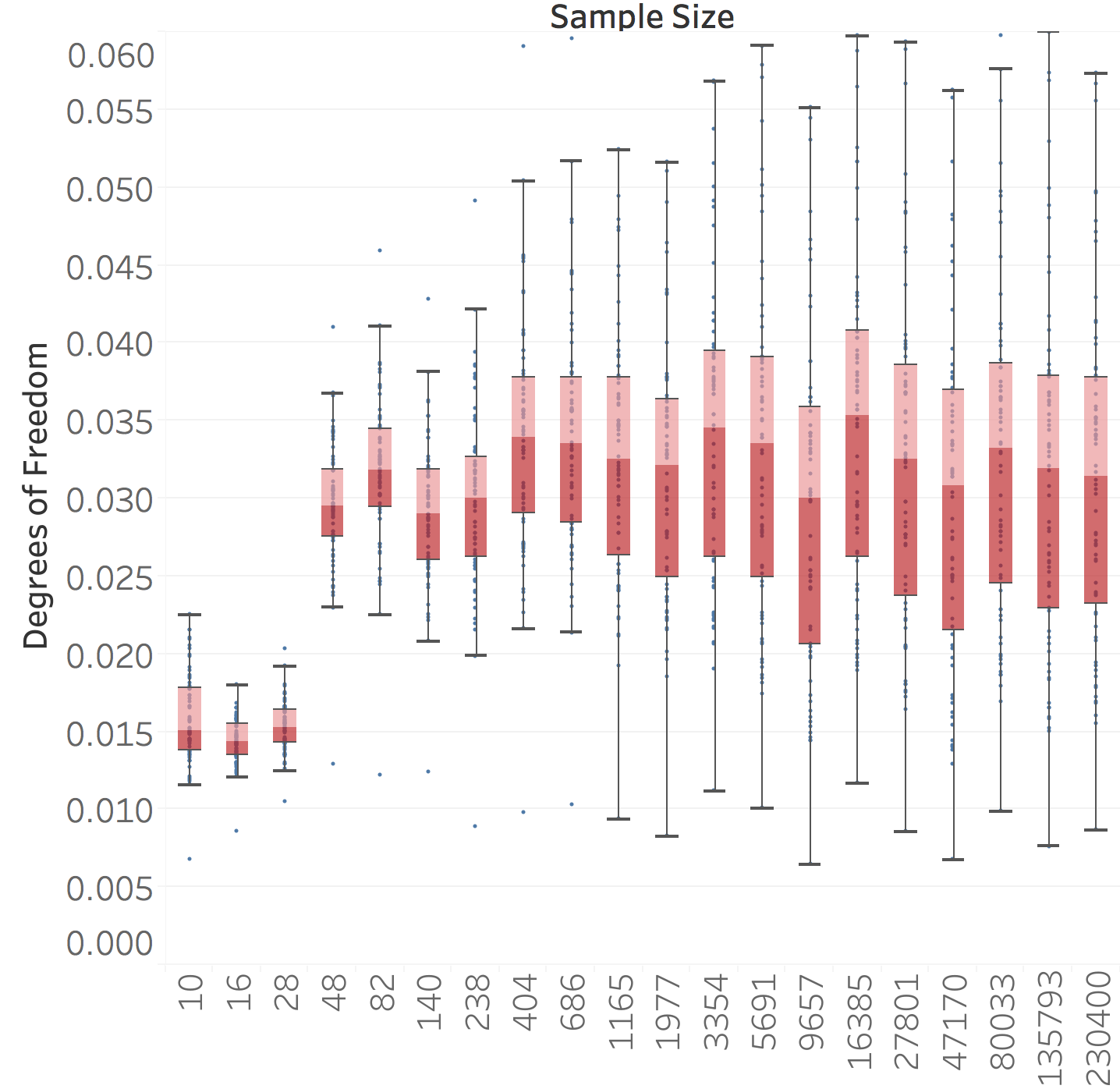}  \\

        \includegraphics[width=0.225\textwidth]{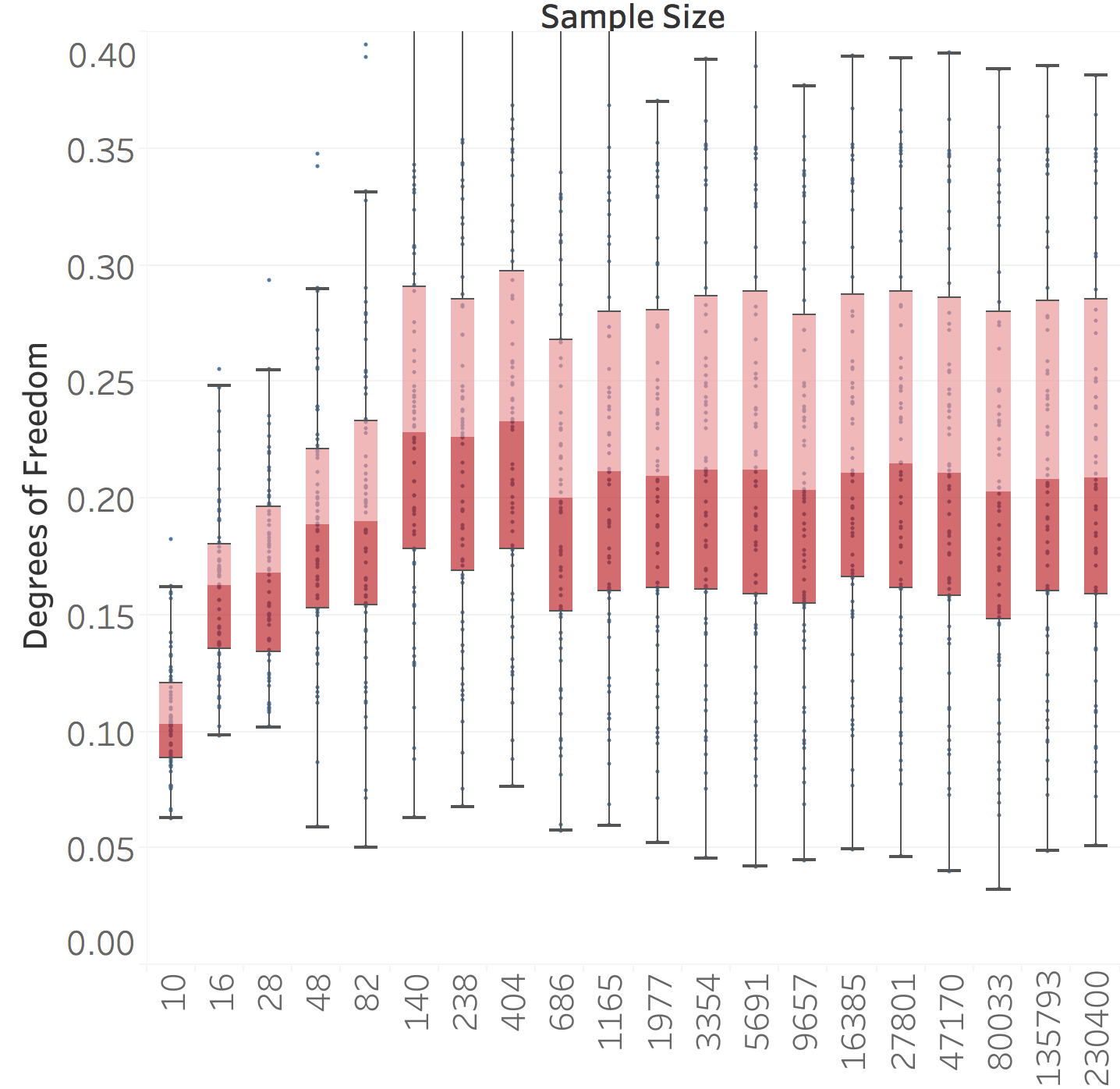} &

        \includegraphics[width=0.225\textwidth]{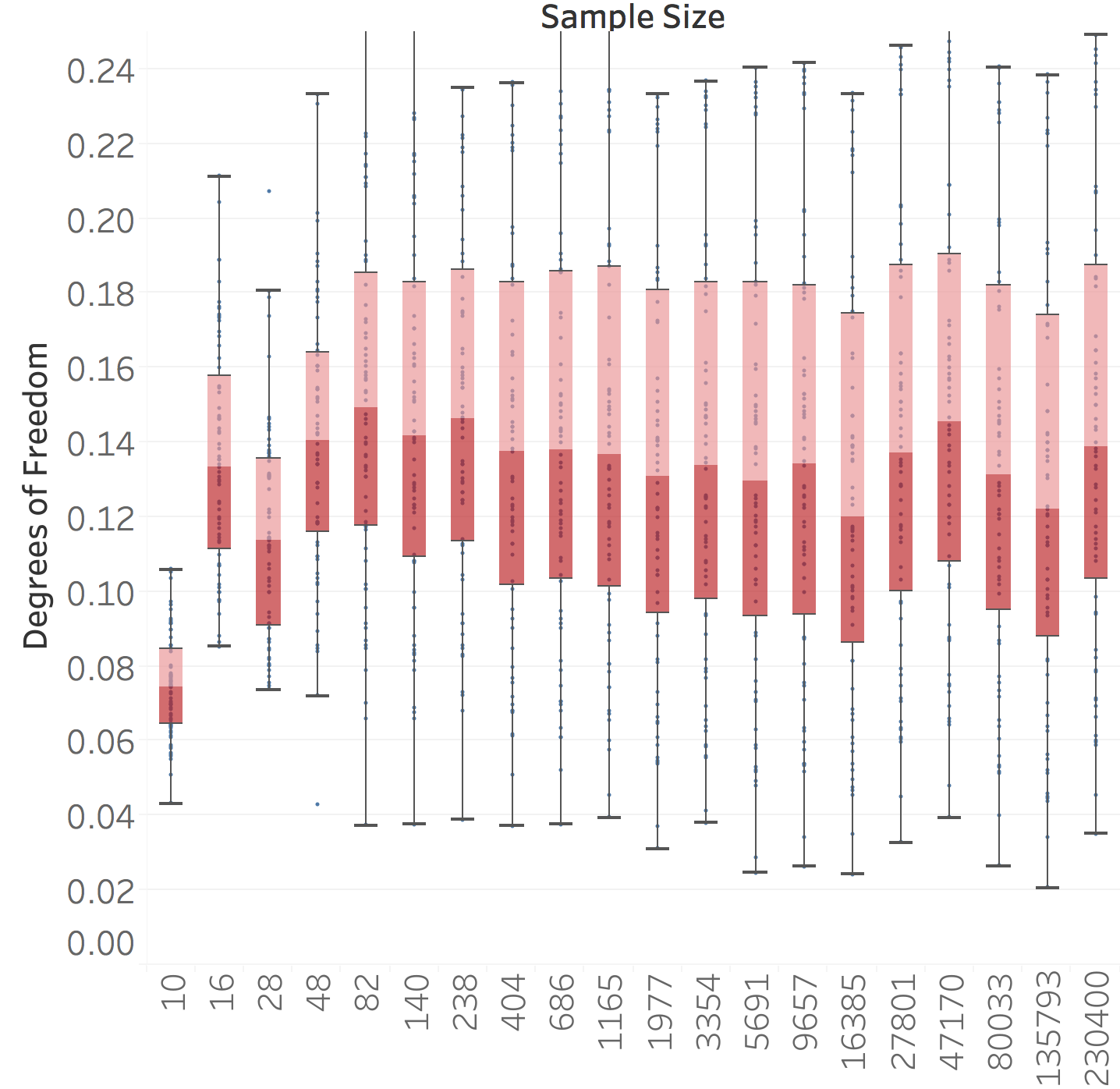}  &

        \includegraphics[width=0.225\textwidth]{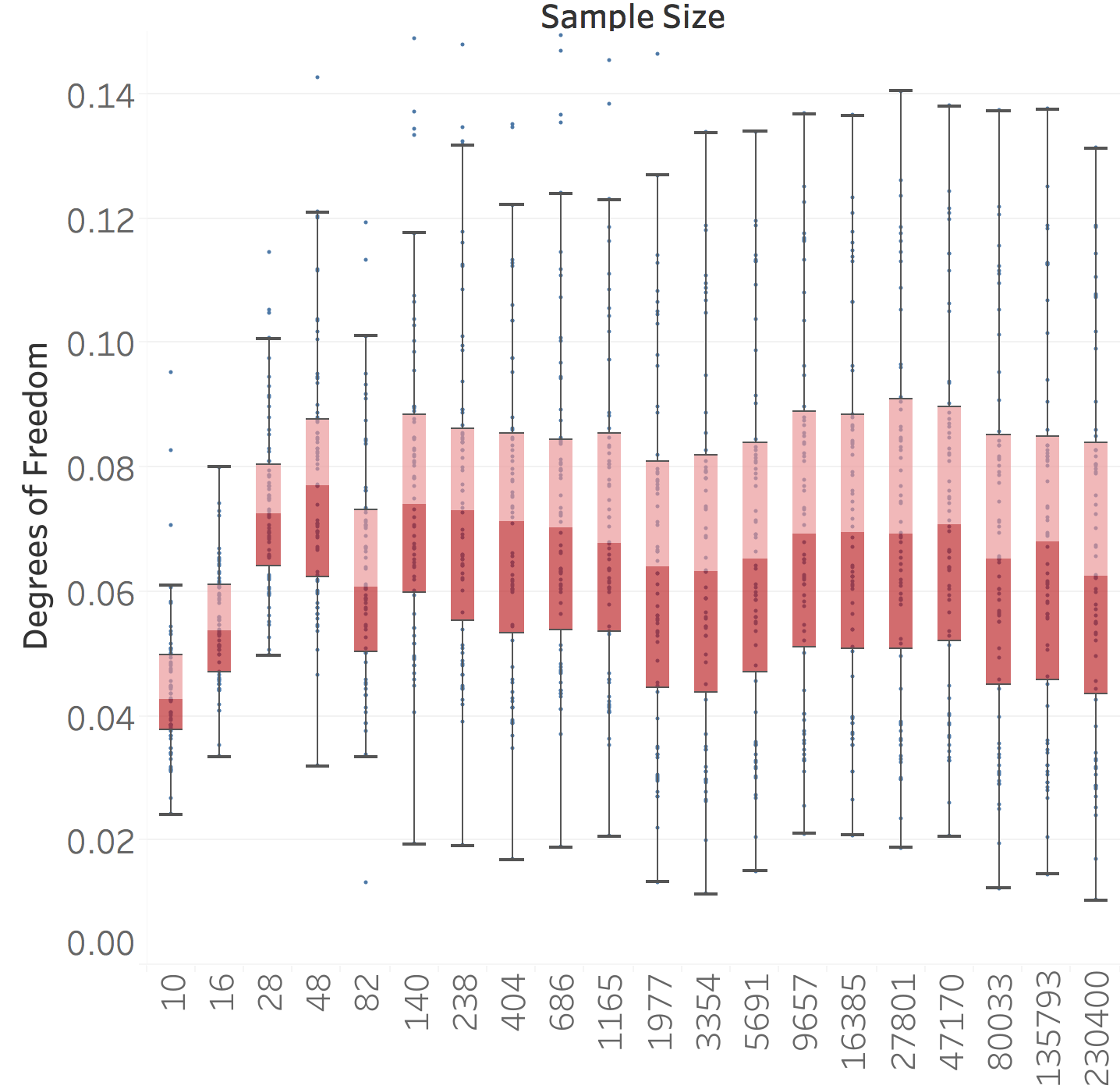}  &

        \includegraphics[width=0.225\textwidth]{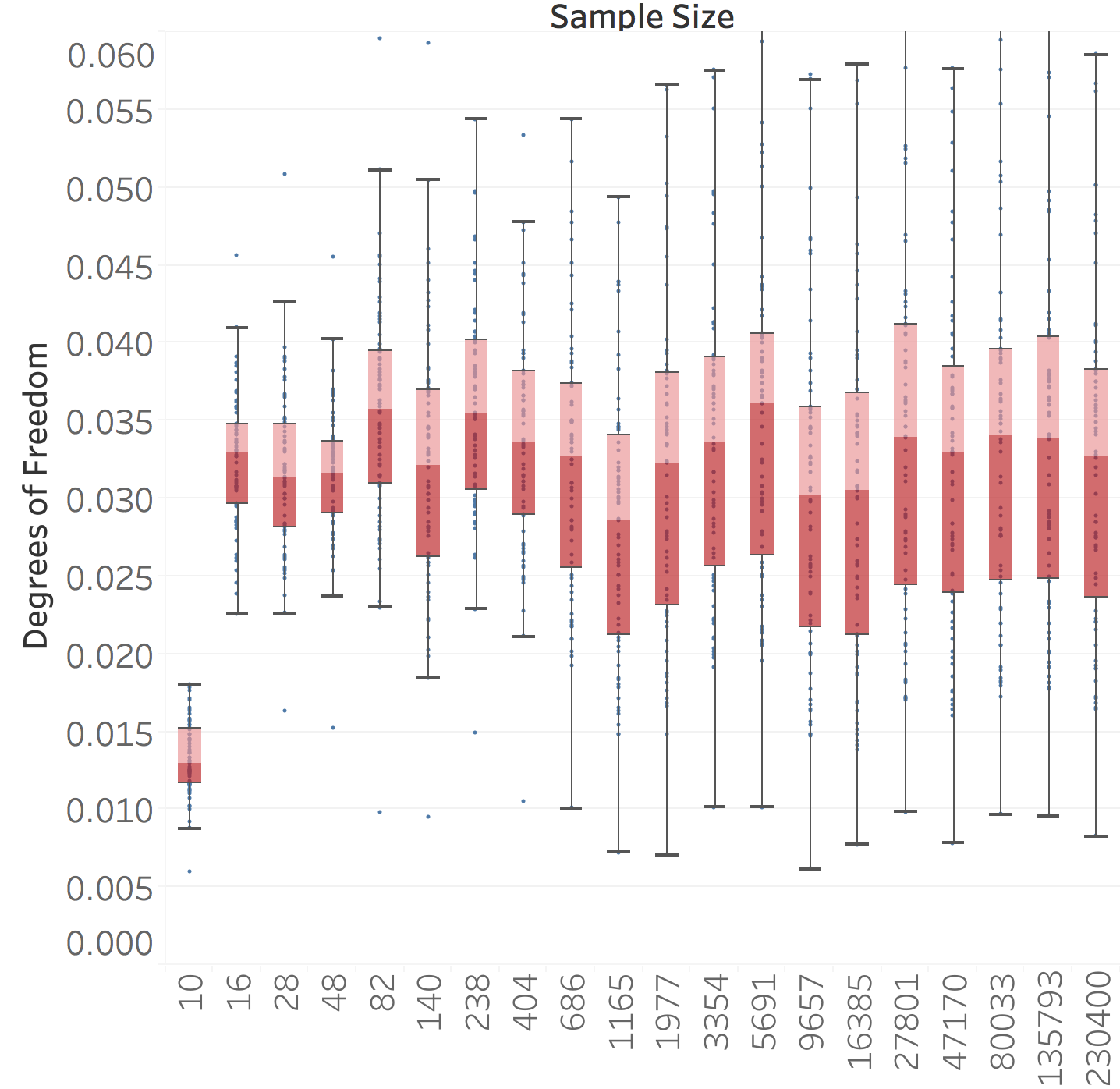} \\  
        \textbf{DOF:} ($\sigma=15$)  & ($\sigma=25$)  & ($\sigma=50$)  & ($\sigma=100$) 
    \end{tabular}
    \caption{\small{\textit{Effects of weight sharing and non-weight sharing and sample size on the RSS and DOF for denoising.} Each column corresponds to a different noise level. The upper row corresponds to WC while the bottom to WS. Each upper panel depicts the normalized RSS $\frac{1}{\sigma^2} \| y - h_{\Theta}(y) \|_2^2$ as a function of train sample size.    Each lower panel depicts the degrees of freedom $\nabla_y \cdot h_{\Theta}(y)$ of as a function of training sample size. The blue dots correspond to $68$ test images on which the RSS was computed. The box and whisker depict the percentiles.}  
    } \label{fig:RSS}
\end{figure*}

Error bars for the individual SURE components including DOF and RSS are also plotted in Fig.~\ref{fig:RSS} , respectively. The upper (res. bottom) rows correspond to WC (res. WS). Fig.~\ref{fig:RSS} depicts the evolution of normalized RSS, namely $\frac{1}{\sigma^2} \| y - h_{\Theta}(y) \|_2^2$ over train sample size. Similarly, Fig.~\ref{fig:RSS} plots the DOF $\nabla_y \cdot h_{\Theta}(y)$. The blue dots correspond to $68$ test image samples. Box-and-whisker plots also depict RSS percentile. It is first observed that for both WS and WC scenaria, DOF (res. RSS) tend to be increasing (res. decreasing) with the train sample size, where it finally saturates at a limiting value that is identical for both WS and WC. Interestingly, the limiting value coincides with the generalization MSE as per \eqref{eq:sure_h}. The DOF for WS scheme, however, ramps off quickly, suggesting that fewer samples are required to construct the bases and attain the degrees of freedom embedded in the network. On the contrary, RSS would drop quickly for WS, which contributes to small SURE values.

Our second observation compares RSS and DOF values for WS and WC. It appears that, under different noise regimes, WS consistently achieves larger DOF. The achieved RSS however is much smaller which renders the overall SURE (or MSE) smaller for WS in low train sample complexity regimes. In addition, upon using sufficient train samples, RSS converges to unity for all noise levels, which corresponds to $\| y - h_{\Theta}(y; \Phi) \|_2^2 = n \sigma^2$. We explain this by noting that any sensible denoising algorithm should output estimated images within the noise level $n \sigma^2$ of the corrupted image.


\vspace{-2mm}
\subsection{Deblurring}
\label{deblur}
\vspace{-2mm}
This section addresses Q2 for natural image deblurring. The sensing matrix $\Phi$ is a linear operator that convoles an image with a Gaussian kernel of standard deviation $1.6$. We prepared data as described for denoising, except that we extracted patches of size $50 \times 50$. The same ResNet as in the denoising case was adopted to model the proximal. However, instead of encouraging data consistency with a gradient step as for denoising (see \ref{eq:gradient_step}), the full least square problem is simply solved after each proximal step. As a result, the state variable update is modified as
\begin{equation}
    s^{t+1} = (\Phi \Phi^{\mathsf{H}} + \alpha I)^{-1} \left( \Phi^{\mathsf{H}} y + \alpha x^t \right).
\end{equation}
It is worth commenting that the approach of tackling a general image restoration problems by repeatedly denoising and solving a least-squares problem is a common practice in image processing; see e.g.,~\cite{romano2017little,zoran2011learning,chan2017plug}.

We train the architecture in the same way described in the previous subsection. The experiments are repeated for two noise levels $\sigma \in \{ \sqrt{2}, 2\}$ at different panels. Each panel depicts the test PSNR of neural proximal algorithm as a function of training sample size. The orange (res. blue) line corresponds to WS (res. WC). We observe a consistent benefit from using weight sharing in sample sizes smaller than $50$K.

\subsection{Compressed sensing MRI}
\label{cs-mri}
To further investigate Q2, we consider also the task of compressed sensing~\cite{donoho2006compressed} for MRI reconstruction. Looking at the linear model (1), compressed sensing (CS) assumes there are typically much less measurements than the unknown image pixels, i.e., $m \ll n$. A prime example for CS is reconstruction of MR images~\cite{pualy_mri20017}, that is widely adopted in the clinical scanners. In essence, the MR scanner acquires a fraction of Fourier coefficients (k-space data) of the underlying image across various coils. We focused on a single-coil MR acquisition model, where for a patient the acquired k-space data admits 
\begin{equation}
y_{i,j} = [\cF(x)]_{i,j},~~(i,j) \in \Omega
\end{equation}
Here, $\cF$ refers to the 2D Fourier transform, and the set $\Omega$ indexes the sampled Fourier coefficients. Just as in conventional CS MRI, we selected $\Omega$ based on variable-density sampling with radial view ordering that is more likely to pick low frequency components from the center of k-space~\cite{pualy_mri20017}. Only $20\%$ of Fourier coefficients were collected.


\noindent\textbf{Dataset.}~It includes $19$ subjects scanned with a 3T GE MR750 whole body MR scanner. Fully sampled sagittal images were acquired with a 3D FSE CUBE sequence with proton density weighting including fat saturation. Other parameters include FOV=$160$mm, TR=$1550$ (sagittal) and $2,000$ (axial), TE=25 (sagittal) and 35 (axial), slice thickness $0.6$mm (sagittal) and 2.5mm (axial). The dataset is publicly available at \cite{mridata.org}. Each image is a complex valued 3D volume
of size $320 \times 320 \times 256$. Axial slices of size $320 \times 256$ were considered as the input for train and test. $16$ patients are used for training ($5,120$ image slices) and 3 patients for test ($960$ image slices).

Neural proximal algorithm with $T=5$ iterations was run with minibatch of size $4$. For any train sample size, training is performed for various learning rates $3 \times \{10^{-6}, 10^{-5}, 10^{-4}, 10^{-3}\}$, choosing the one achieving the highest PSNR. The input and output were complex-valued images of the same size and each included two channels for real and imaginary components. The input image $x_0=\Phi^{\mathsf{H}}y$ was indeed the inverse 2D FFT of the k-space data where the missing frequencies were filled with zeros. It was thus severely contaminated with aliasing artifacts. The benefits of weight sharing for small sample sizes is evident, when using only a few MR images for training would lead to around $1$ dB gain compared with the weight changing scheme. The gap kept decreasing with the train sample size and finally after $10^2$ samples the weight changing scheme led to an slight improvement, possibly due to the larger representation capacity. Note also that compared with the denoising and deblurring experiments, the gap disappears for a smaller sample sizes, as train images are of large dimension with $320 \times 256$ training pixels.

\begin{figure}
\centering
\begin{tabular}{c}
     \centering
     \includegraphics[width=0.75\textwidth]{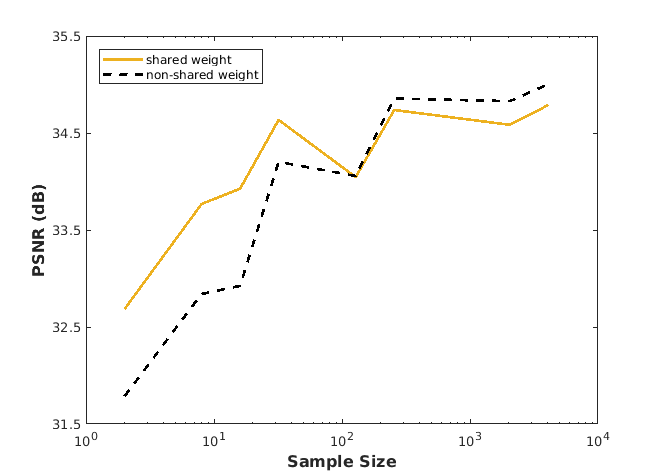} 
\end{tabular}

    \caption{\textit{Effects of weight sharing and sample size on the performance for CS-MRI with $5$-fold undersampling of the $k$-space data. The panel depicts the test PSNR as a function of training sample size. The orange line corresponds to the result obtained with WS and the blue line to WC.}  } \label{fig:cs-mri}
\end{figure}

\subsection{Filtering interpretation of proximals}
\label{scattering-nets}
\vspace{-2mm}

\begin{figure*}[t]
    \centering
    \begin{tabular}{c c c c c}
        \centering
        \includegraphics[width=0.125\textwidth]{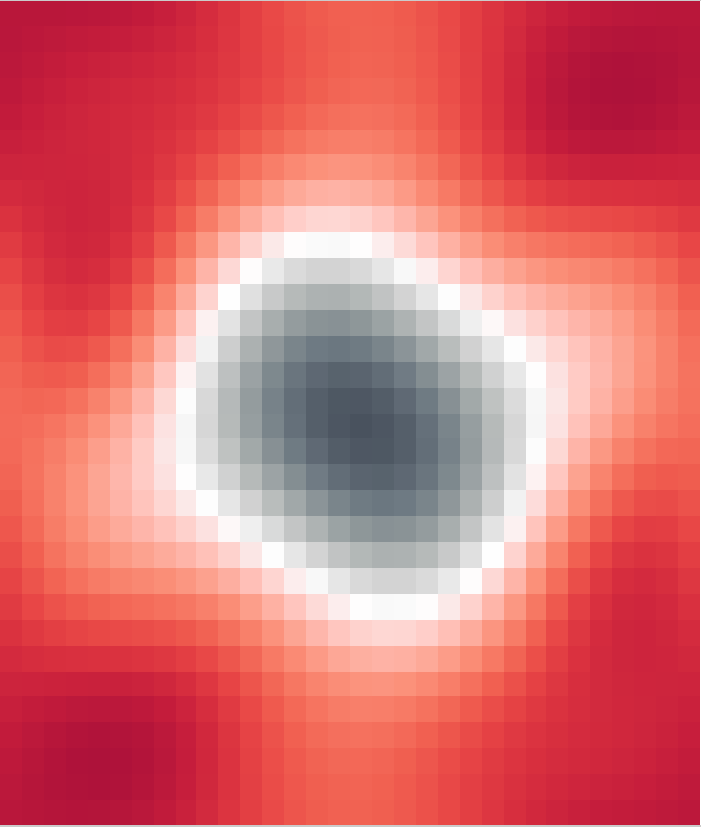} &

        \includegraphics[width=0.125\textwidth]{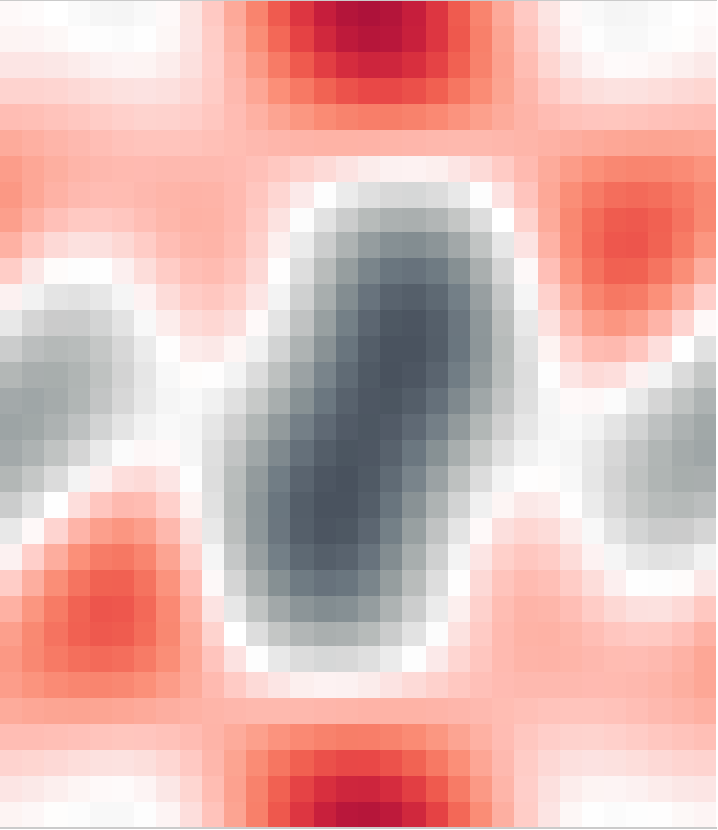} &

        \includegraphics[width=0.125\textwidth]{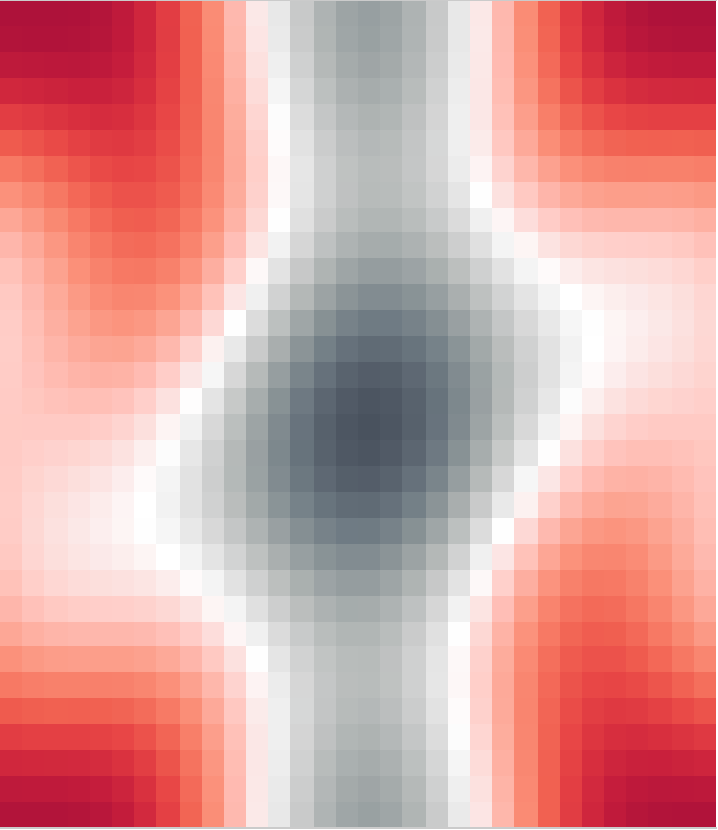}  &

        \includegraphics[width=0.125\textwidth]{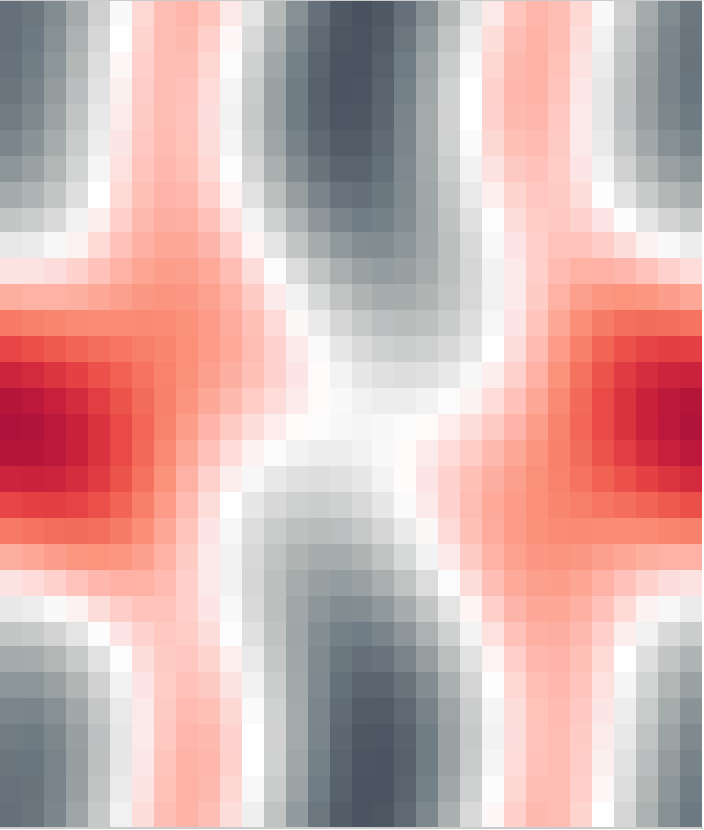}  &
          

        \includegraphics[width=0.125\textwidth]{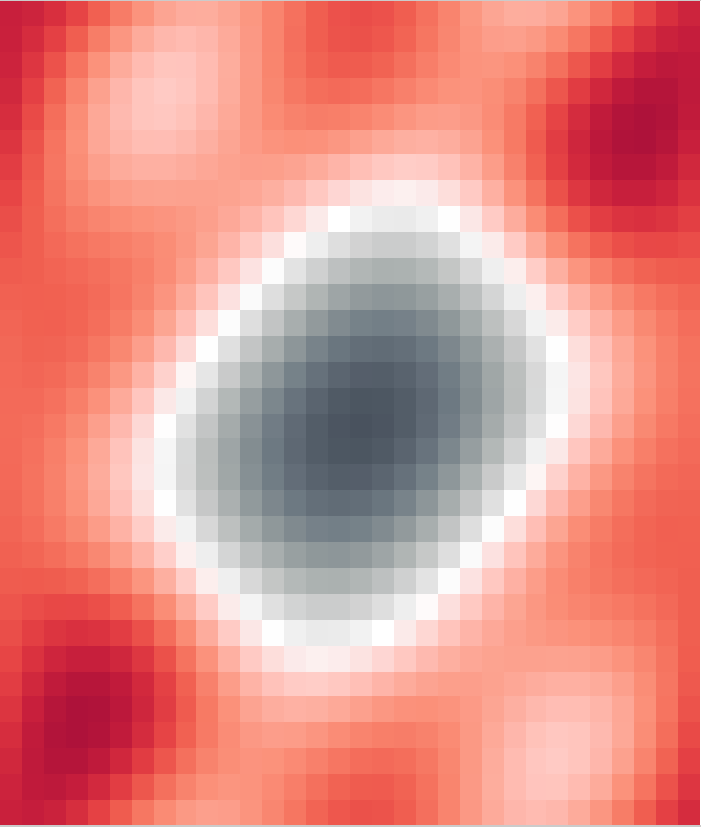}  \\


        \includegraphics[width=0.125\textwidth]{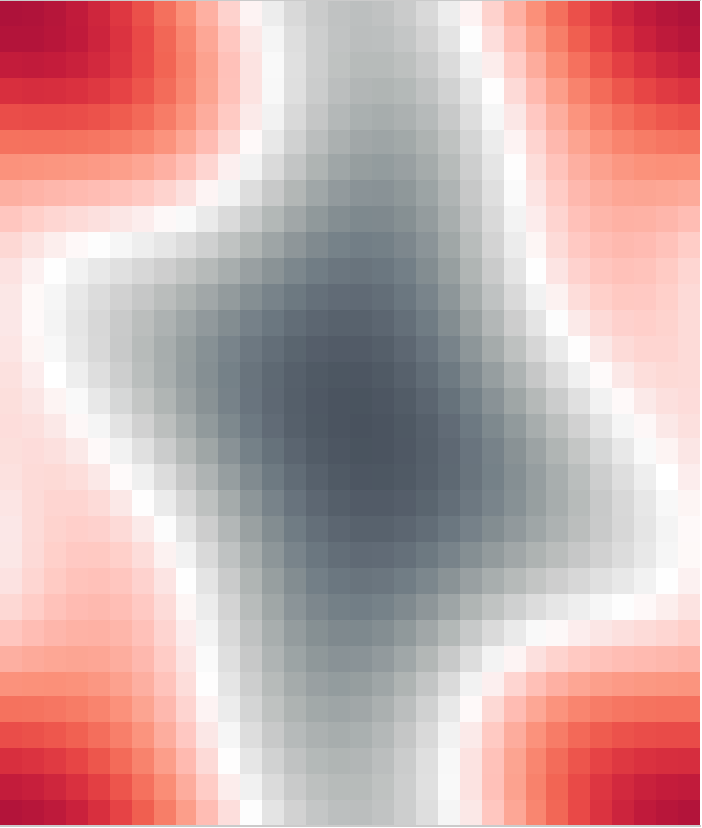}   &

        \includegraphics[width=0.125\textwidth]{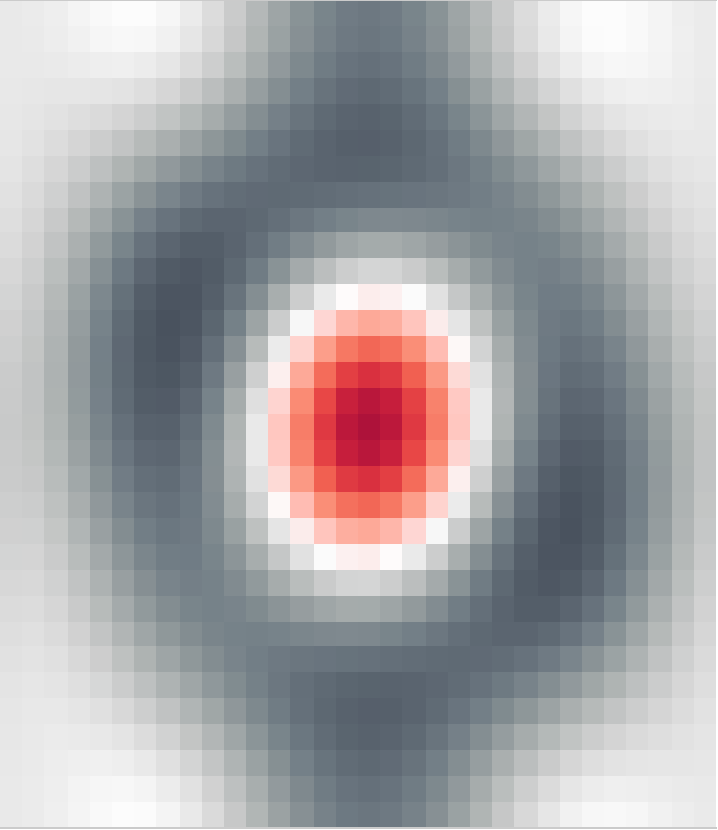}   &

        \includegraphics[width=0.125\textwidth]{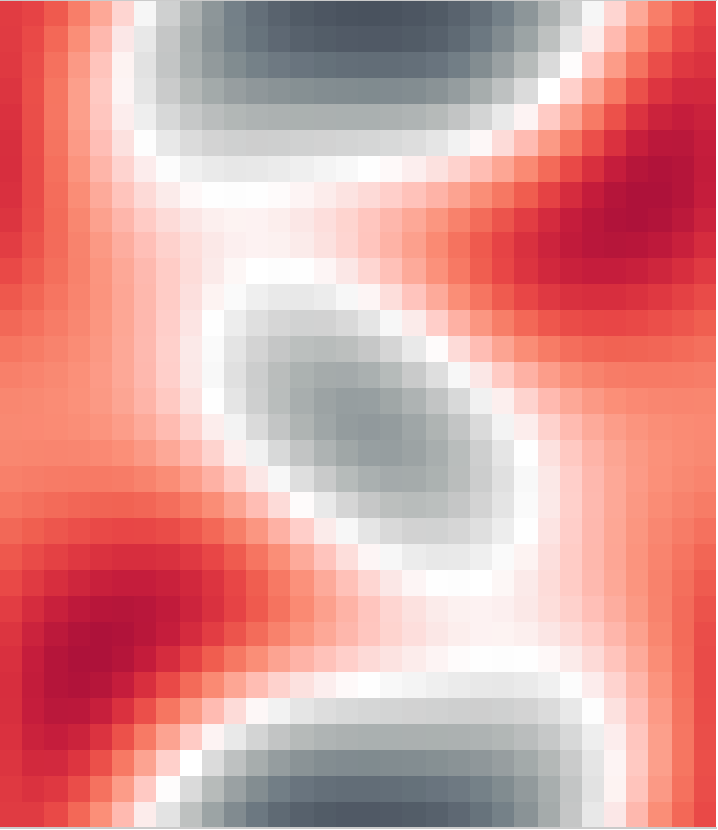}   &

        \includegraphics[width=0.125\textwidth]{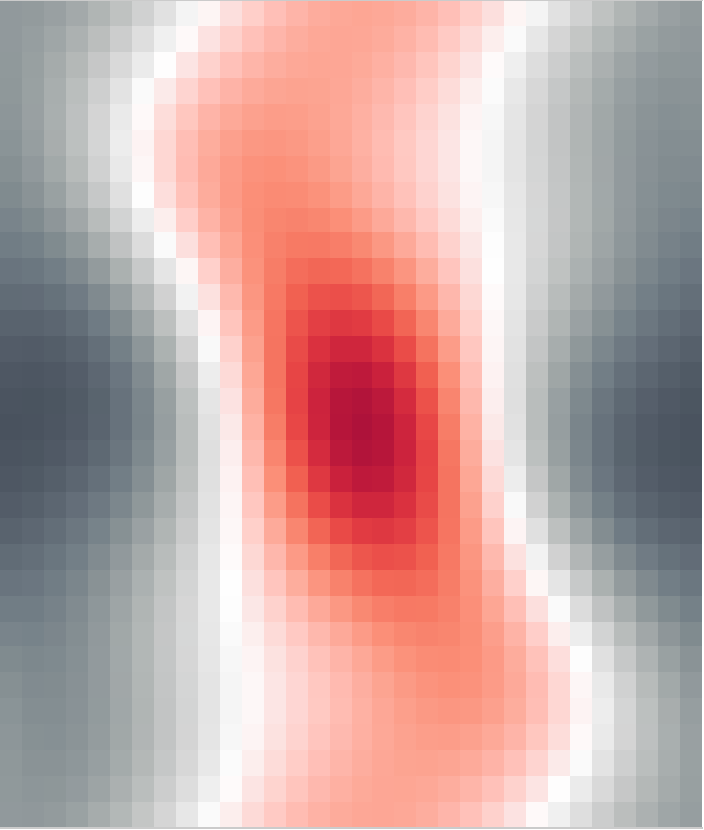}  &

        \includegraphics[width=0.125\textwidth]{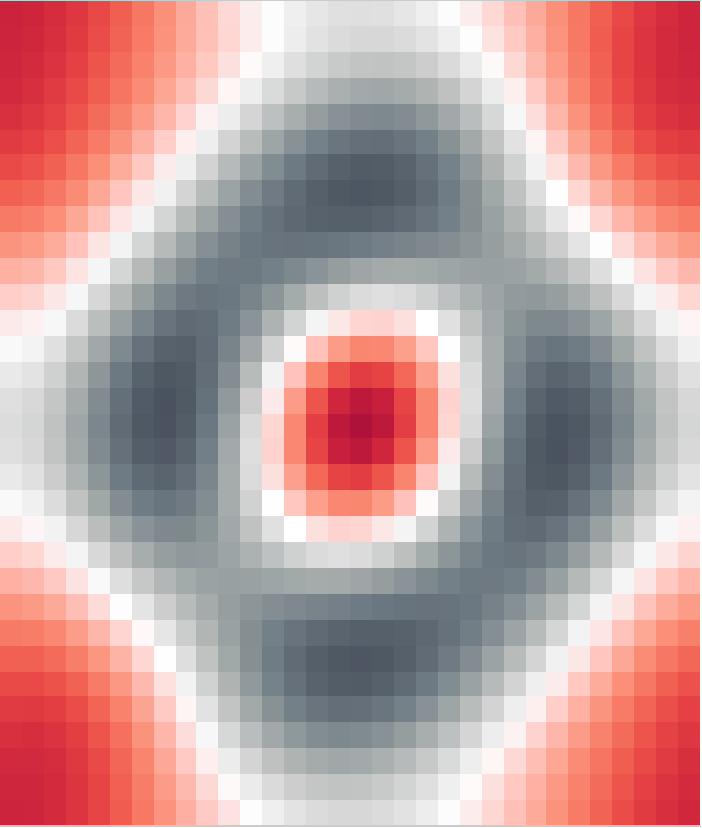}  \\


        \includegraphics[width=0.125\textwidth]{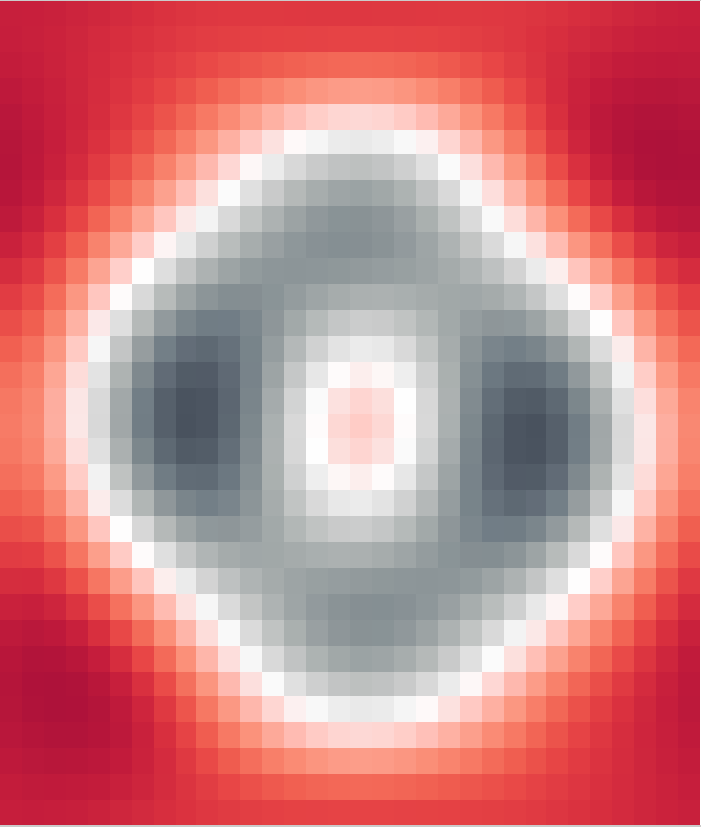}    &

        \includegraphics[width=0.125\textwidth]{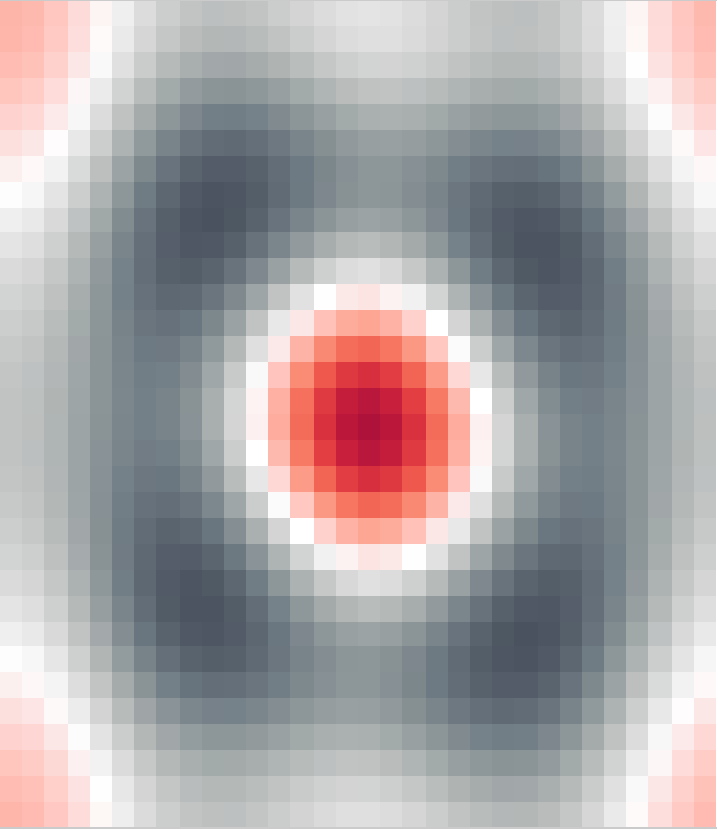}    &

        \includegraphics[width=0.125\textwidth]{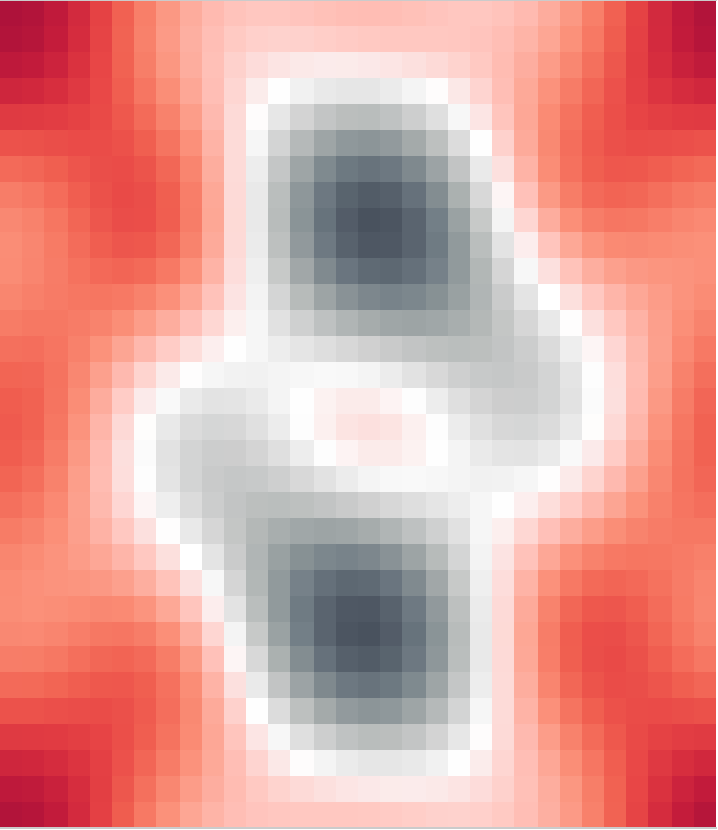}   &

        \includegraphics[width=0.125\textwidth]{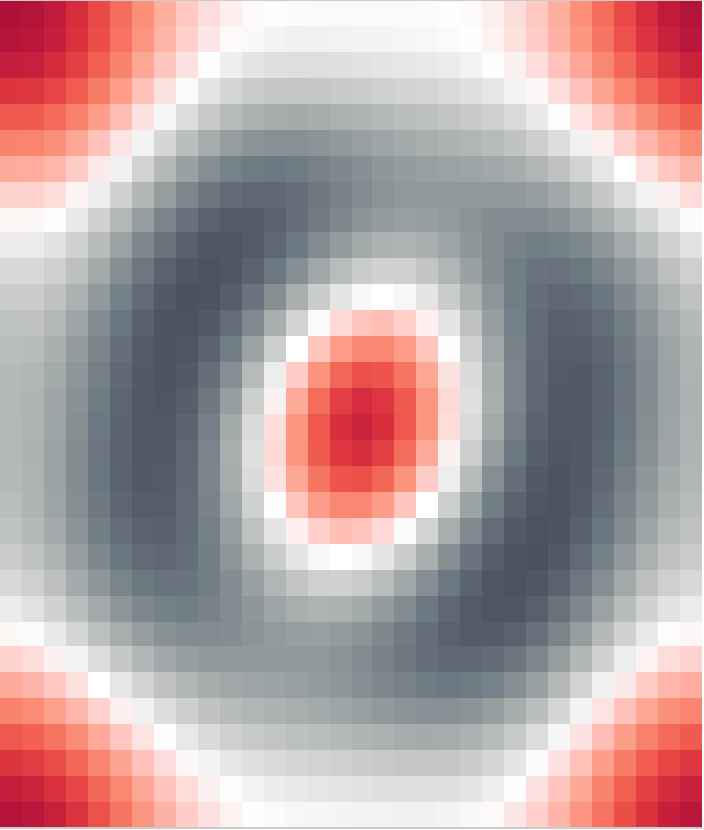}  &

        \includegraphics[width=0.125\textwidth]{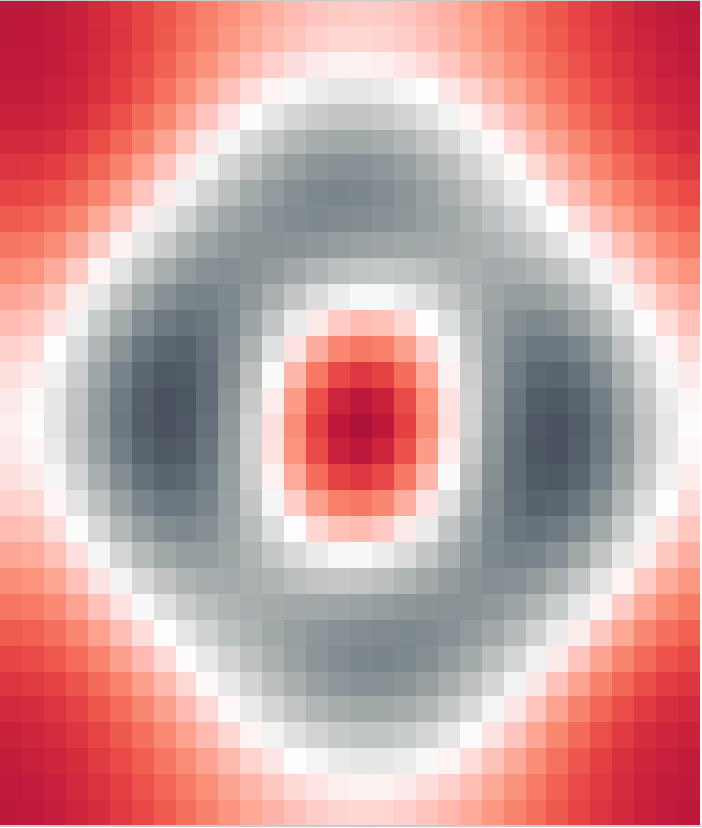}   \\


        \includegraphics[width=0.125\textwidth]{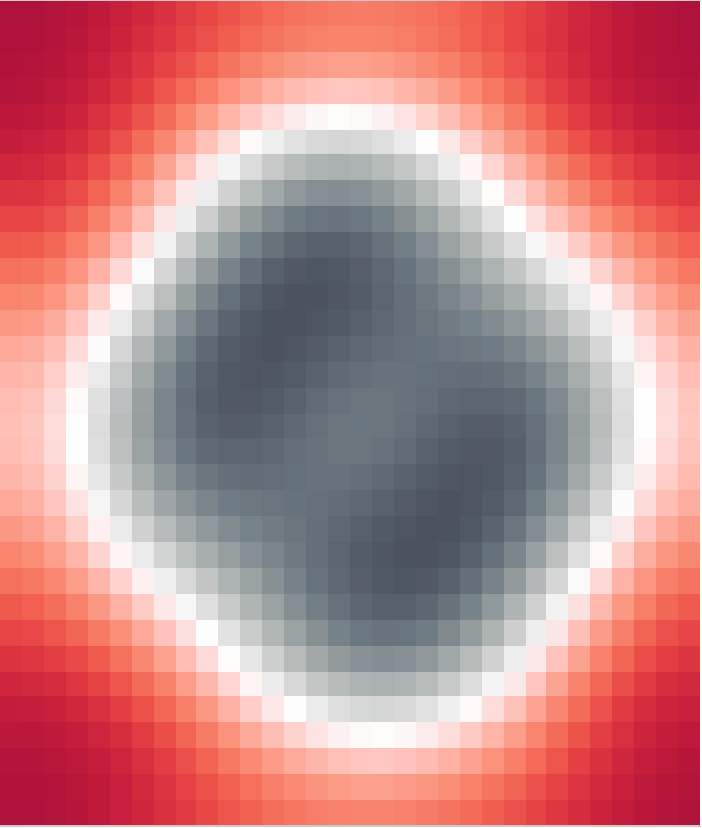}   &

        \includegraphics[width=0.125\textwidth]{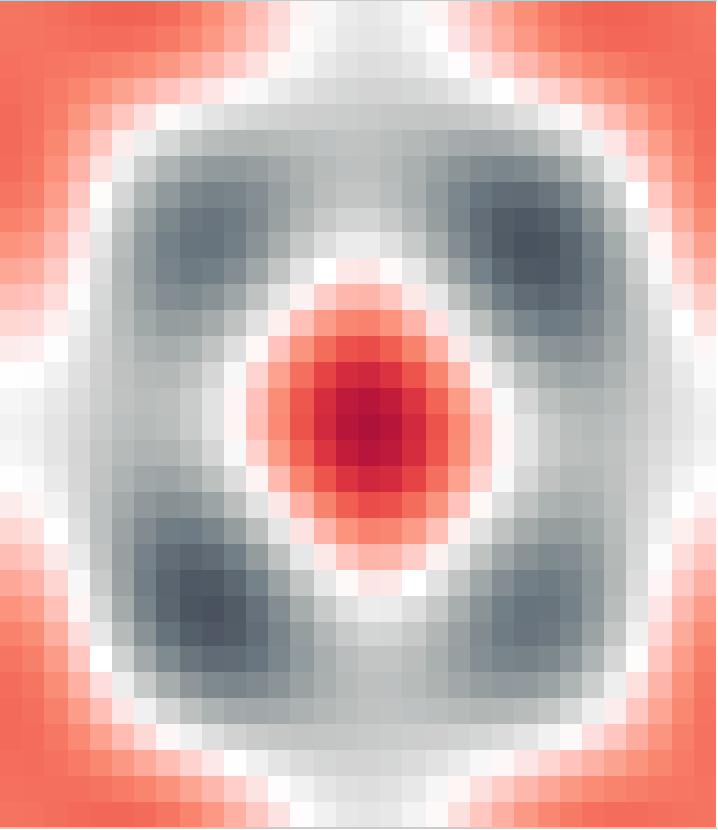}  &

        \includegraphics[width=0.125\textwidth]{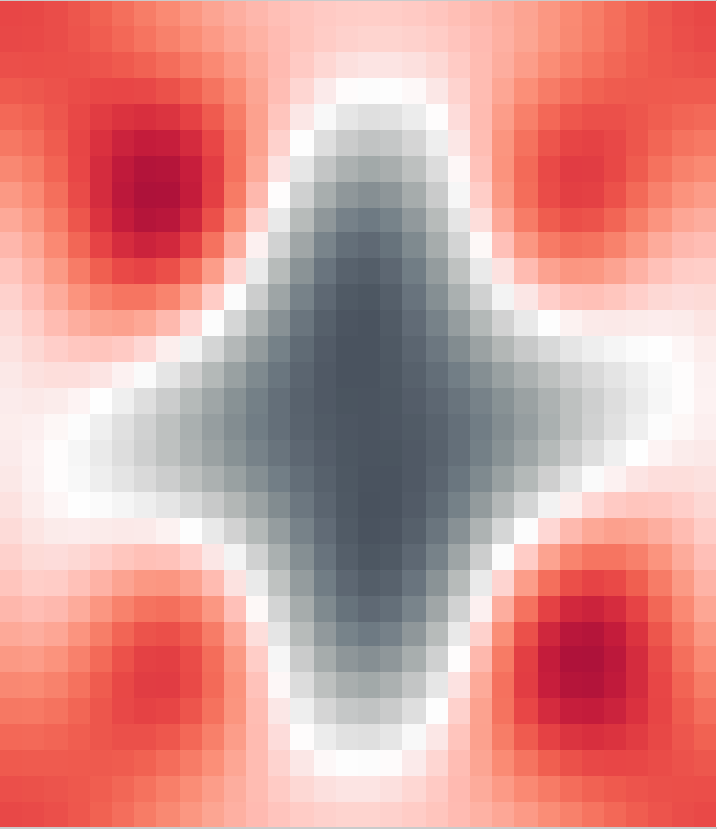}   &

        \includegraphics[width=0.125\textwidth]{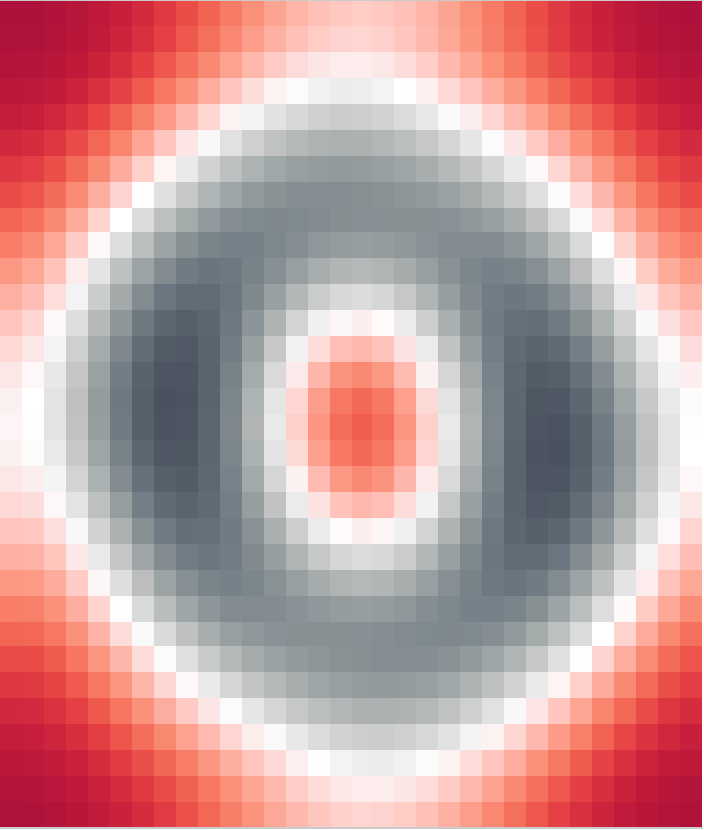}  &

        \includegraphics[width=0.125\textwidth]{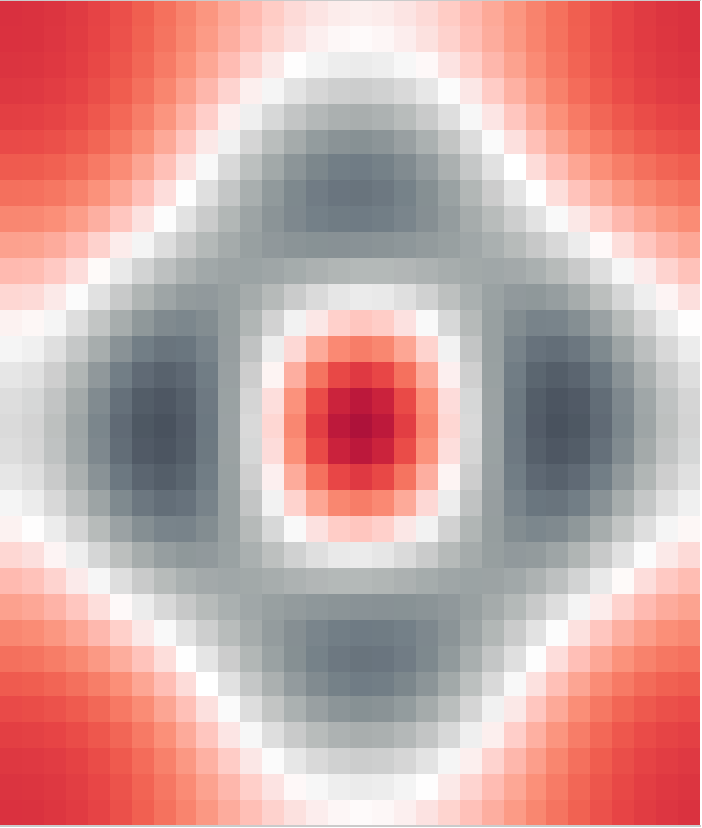} \\  
        
    \end{tabular}
    
    \caption{\textit{Frequency visualization of the learned proximals in the shared and non-shared case.} The first four columns correspond to the weights obtained in the $4$ iterations of the non-shared weights. The fifth column corresponds to the shared-weights case. Rows from top to bottom are for $10;686;5,691;135,793$ train samples. Each panel depicts the summation of the magnitudes of the 2D Fourier transforms of the filters. The noise level is fixed to $\sigma=100$. The color coding associates large magnitudes to black, and small ones to red. } \label{fig:weights}
\end{figure*}

Along with Q3, it is of interest to \textit{explain} how the cascade of learned proximals of WS and WC contribute to recover the input image from errors. To do so, we focus on the natural image denoising described in Section 4.2 with $T=4$ proximal iterations. Recall the $t$-th proximal network mapping $s_{t+1}$ to $x_{t+1}$ through a ResNet with 2RBs. We focus on the first convolutional layer with $128$ kernels collected as $\{f_{t,i}\}_{i=1}^{128}$ per iteration $t$. For visualization purposes, we propose to compute the two-dimensional Fourier transform for each of filters $\{f_{t,i}\}_{i=1}^{128}$ and then to sum over all filters the magnitude of the Fourier coefficients. We repeat this process for $T$ iterations in WC case and the single set of filters in WS.

The results are shown in Fig.~\ref{fig:weights} for the noise level $\sigma=100$. The first four panel columns represent the weights obtained in four iterations of WC network, while the fifth column represents WS. Each row corresponds to a different sample size. It is observed that for WS at high sample sizes the filters converge to a spectrum associated with a bandpass filter. The pattern observed for WC however is interesting; the odd iterations converge to a lowpasss, while the even iterations converge to bandpass filters. This is reminiscent of the scattering transform~\cite{bruna2013invariant,anden2014deep}, where a cascade of lowpasss and bandpass operations is applied.

In contrast with scattering networks however, our \textit{learned} proximal network applies lowpasss filtering followed by bandpass filtering. The scattering transform applies several lowpasss filters, and then a highpasss filter at the last layer. We also observe that the shared weights converge to the final spectrum at approximately $686$ examples, while the non-shared case requires more than $5,691$ samples to converge. Moreover, for WC case the filters in the first two iterations are very similar to their final versions already at $686$ examples, meaning that the filters in the first iterations are trained first.

\vspace{-2mm}
\section{Conclusions}
\label{conc}
\vspace{-2mm}
This paper investigates the generalization risk of unrolled neural networks appearing in image translation tasks. The Stein's Unbiased Risk Estimator (SURE) is adopted as an estimate for the generalization risk, and the DOF term quantifying the prediction variance that is the trace of the end-to-end network Jacobian is analyzed. Under certain incoherence conditions on the train weight matrices, DOF is derived in terms of the weight path sparsity of the network. Extensive empirical evaluations are performed with natural images for image denoising and deblurring tasks. While the analysis are performed for the recurrent unrolled networks, non-recurrent networks are empirically tested for comparison. The observations indicate that the DOF  increases  with  train  sample  size and  converges  to  the  generalization  risk  for  both  recurrent  and  non-recurrent schemes. In addition, the recurrent network converges significantly quicker (with less train samples) compared with non-recurrent scheme, hence recurrent scheme serves as a regularization for low sample size regimes.  All in all, this is the first attempt to apply SURE for generalization risk analysis of unrolled neural networks.


There are still important avenues to explore that are left for future research. One such avenue pertains to extending the SURE analysis to arbitrary sensing matrices. Another one includes understanding the link between early stopping and weight sharing for training unrolled neural networks.


\section{Appendix}
\subsection{Proof of Lemma 2}
\label{proof_lemma2}

Upon defining the sample correlation matrix $C_x^N := \frac{1}{N} \sum_{i=1}^N x_i x_i^{\mathsf{H}}$, the network weights come from the training process that optimizes    
\begin{align*}
W & = \arg\min_{W} \frac{1}{N}\sum_{i=1}^N \|x_i - (I-\cP_{W}) y_i\|^2 \\
     &  \overset{(a)}{\approx} \arg\min_{W} \frac{1}{N} \sum_{i=1}^N \|\cP_{W}x_i\|^2 + \hat{\sigma}^2 \tr((I-\cP_{W})) \\
     & = \arg\min_{W} \tr\big(\cP_{W} (C_x - \hat{\sigma}^2 I) \big) \label{eq:train_cost_linear}
\end{align*}
where the approximation (a) comes from $\frac{1}{N} \sum_{i=1}^N v_i v_i^{\mathsf{H}} \approx \hat{\sigma}^2 I$, and quickly approaches $\sigma^2 I$ for relatively large $N$. 

Apparently, the training objective amounts to learning the bases for principal component analysis (PCA). The training process then tunes the network weights to the singular vectors of the sample correlation matrix. Let us decompose the sample correlation matrix as $C_x^N=U\Sigma U^{\mathsf{H}}$. In essence, the optimal $W \in \mathbb{R}^{\ell \times n}$ is orthogonal with the rows that include the singular vectors $\{u_i\}$ where $\sigma_i^N \leq \hat{\sigma}^2$; if $\sigma_i^N \leq \hat{\sigma}^2$, we set the corresponding row to zero. For the end-to-end map $J_{\infty}=I - \cP_{W}$ the DOF is then reduced to $n- \tr(\cP_W)$, and the result of Lemma 2 immediately follows.


\subsection{Proof of Lemma 4}
\label{proof_lemma3}
%






Let 
\begin{equation}
    Q= W W^{\mathsf{H}}-\mbox{diag}(b)
\end{equation}
then the diagonal of $ Q$ are all zeros, therefore  
\begin{equation}
   \tr( D_{i} Q ) = 0. 
\end{equation}
Let $\lambda_j(D_{i} Q)$ be eigenvalues of $D_{i} Q$, we rank the eigenvalues so that $\lambda_1(D_{i} Q)$ be the eigenvalue with largest absolute value, then $|\lambda_1(D_{i} Q)|$ is the spectral norm of the matrix as $\|D_{i} Q \|_2$.

Now we find upper bound on the spectral norm. 
Using Gershgorin circle theorem, there exist at least one index $k$ such that
\begin{equation}
 |\lambda_1(D_{i}  Q ) |  \le \sum_{j \ne k} |(D_{i}  Q ))_{kj}|   
\end{equation}
Since the diagonal entry of $D_{i}  Q $ is zero, so
\begin{align}
\EE\|D_{i} Q \|_2 &= \EE |\lambda_1(Q)|\\
  &\leq 
  \EE[
  \sum_{j\neq 1}|[D_{i} Q]_{1j}|]
   \\
&\leq
(s_i-1)\mu_{W} 
\end{align}

Now we could bound all the following quantities using spectral norm, 
\[ 
\begin{array}{ll}
  \|D_{i} Q \|_F^2
 &=  \tr( D_{i} Q Q^{\mathsf{H}} D_{i_j}^{\mathsf{H}})\\
  &= |\sum_{j=1}^{s_i} \lambda_j(D_{i}  Q )^2|
\\
&\leq s_i \|D_{i} Q \|_2^2.
\end{array}
\]
Using inequality for trace of products of matrices~\cite{shebrawi2013trace}, we have a bound
\[ 
\begin{array}{ll}
  |\tr( D_{i_j} Q \ldots  Q  D_{i_{1}}  Q )|
&\leq \prod_{l=1}^j \|D_{i_l} Q \|_F \\
&\leq  \prod_{l=1}^j\sqrt{s_{i_l}}  \|D_{i} Q \|_2
\end{array}
\]



\subsection{Proof of Theorem 1}
Plugging the result of Lemma 4 into to the Jacobian expansion in (10) and (11), we have the decomposition of Jacobian as follows
\begin{align}
 \EE \tr[J_{T}]  &=  \EE \big[ \tr \big( \prod_{\tau=0}^{T} (I- W^{\mathsf{H}} D_{i} W) \big) \big]\\
 &= \EE \big[ \sum_{j=0}^T \sum_{i_1, \ldots, i_j} (-1)^{j} \tr\big(W^{\mathsf{H}} D_{i_j} W W^{\mathsf{H}} \ldots W W^{\mathsf{H}} D_{i_{1}} W \big) \big] \\
 &= \EE \big[\tr(I) -\sum_{i=0}^{T}\tr( D_{T-\tau} WW^{\mathsf{H}}) + \ldots + \sum_{i_1, \ldots, i_T} (-1)^{T} \tr\big(W^{\mathsf{H}} D_{i_T} W W^{\mathsf{H}} \ldots W W^{\mathsf{H}} D_{i_{1}} W \big) \big]
\end{align}
After some rearrangements the deviation bound from the Lemma 4 would result in

\begin{align}
 & \Big| \EE[\tr(J_{T})] - \big(n+\sum_{\cI} (-1)^{|\cI|} p_{\cI} \big) \Big| \\
& \leq \sum \Big| \EE \big[ \tr\big(W^{\mathsf{H}} D_{i_j} W W^{\mathsf{H}} \ldots W W^{\mathsf{H}} D_{i_{1}} W \big) \big]
-     p_{\cI} \Big| 
\\
& \leq  \sum  \prod_{l=1}^j [\sqrt{s_{i_l}} (s_{i_l}-1) \mu_{W} ]
\\
& \leq  {T \choose 2} \epsilon^2 + \ldots + {T \choose T}
\\
& = (1 +  \epsilon)^T - 1 - \epsilon T. 
\end{align}

\newpage 
\bibliography{neurips_paper}
\bibliographystyle{unsrt}

\end{document}